\documentclass[letterpaper, 10 pt, conference]{ieeeconf}  

\IEEEoverridecommandlockouts                              

\usepackage{graphicx}
\usepackage{epstopdf}
\usepackage{epsfig} 
\usepackage{amsmath} 
\usepackage{amssymb}  
\usepackage{mathtools}
\usepackage{amsfonts}
\usepackage{url}
\usepackage[]{units}
\usepackage{tikz}
\usepackage{color}				
\usepackage{transparent}		
\usepackage{lipsum}
\usepackage{multicol}

\usetikzlibrary{calc}
\usetikzlibrary{positioning}
\usetikzlibrary{arrows.meta}
\usetikzlibrary{shadings,intersections}

\usepackage[utf8]{inputenc}
\usepackage{pgfplots}
\pgfplotsset{compat=newest}
\usepgfplotslibrary{groupplots}
\usepgfplotslibrary{dateplot}

\newcommand{\pythonfigwidth}{0.52}

\newcommand{\spacetofig}{-15pt}
\newcommand{\spacetocaption}{-7pt}

\DeclareMathSizes{10}{10}{6}{1}

\title{\LARGE \bf
A Fast and Reliable Pick-and-Place Application with \\a Spherical Soft Robotic Arm
}

\author{Jasan Zughaibi, Matthias Hofer, and Raffaello D'Andrea
\thanks{The authors are members of the Institute for Dynamic Systems and Control, ETH Zurich, Switzerland.
{\tt\small\{zjasan, hofermat, rdandrea\}@ethz.ch}}%
}

\begin{document}

\maketitle
\thispagestyle{empty}
\pagestyle{empty}

\begin{abstract}\label{sec:Abstract}
This paper presents the application of a learning control approach for the realization of a fast and reliable pick-and-place application with a spherical soft robotic arm. The arm is characterized by a lightweight design and exhibits compliant behavior due to the soft materials deployed. A soft, continuum joint is employed, which allows for simultaneous control of one translational and two rotational degrees of freedom in a single joint. This allows us to axially approach and pick an object with the attached suction cup during the pick-and-place application. A control allocation based on pressure differences and the antagonistic actuator configuration is introduced, allowing decoupling of the system dynamics and simplifying the modeling and control. A linear parameter-varying model is identified, which is parametrized by the attached load mass and a parameter related to the joint stiffness. A gain-scheduled feedback controller is proposed, which asymptotically stabilizes the robotic system for aggressive tuning and over large variations of the parameters considered. The control architecture is augmented with an iterative learning control scheme enabling accurate tracking of aggressive trajectories involving set point transitions of 60 degrees within 0.3 seconds (no mass attached) to 0.6 seconds (load mass attached). The modeling and control approach proposed results in a reliable realization of a pick-and-place application and is experimentally demonstrated.   
\end{abstract}


\section{Introduction}\label{sec:Introduction}

Pneumatically actuated soft robotic systems show promise for a variety of applications due to their intrinsic properties \cite{rus2015overview}, \cite{polygerinos_overview}, \cite{gaiser2012overview}. In particular, the inherent compliance and the low weight reduce the risk of injury in case of an uncontrolled impact, allowing for close human-robot collaboration \cite{sanan_pHRI_soft}, \cite{abidi_onIntrinsicSafety}. The combination of low inertia and pneumatic actuation enables fast actuation as shown in \cite{polygerRapidSoft} and \cite{mhofer:ILCaggr}. 

By arranging multiple actuators antagonistically around a joint, both stiffness and position can be controlled independently \cite{althoefer2018antagonistic}. The simultaneous position and stiffness control of a fully inflatable soft robotic arm with a single degree of freedom is presented in \cite{gillespie:simPosStiffCtrl}. An approach to adjust the stiffness of a spherical robotic arm combining two degrees of freedom in a single joint is presented in \cite{mhofer:DesFabModCtrl_V3}. 

Pneumatically actuated soft robotic systems often exhibit complex dynamics due to the viscoelastic material behavior, making the accurate control of such systems challenging \cite{polygerinos_overview}. This particularly applies for the reference tracking of aggressive maneuvers. For the purpose of feedback controller synthesis, it is a common strategy to rely on low-complexity models and compensate for uncertainty by feedback. A linear time-invariant model is identified in \cite{mhofer:ILCaggr} and used in an Iterative Learning Control (ILC) scheme to ensure accurate tracking for aggressive trajectories. The authors of \cite{LPV_modCtrlHinf} synthesize a gain-scheduled feedback controller for reference tracking based on a Linear Parameter-Varying (LPV) system. The control system is further augmented with an ILC scheme. The application of ILC to improve the tracking performance and preserve the compliance of the system at the same time is presented in \cite{angeliniILCsoft_decentra}.  
\begin{figure}
	\centering
	\includegraphics[scale=0.18]{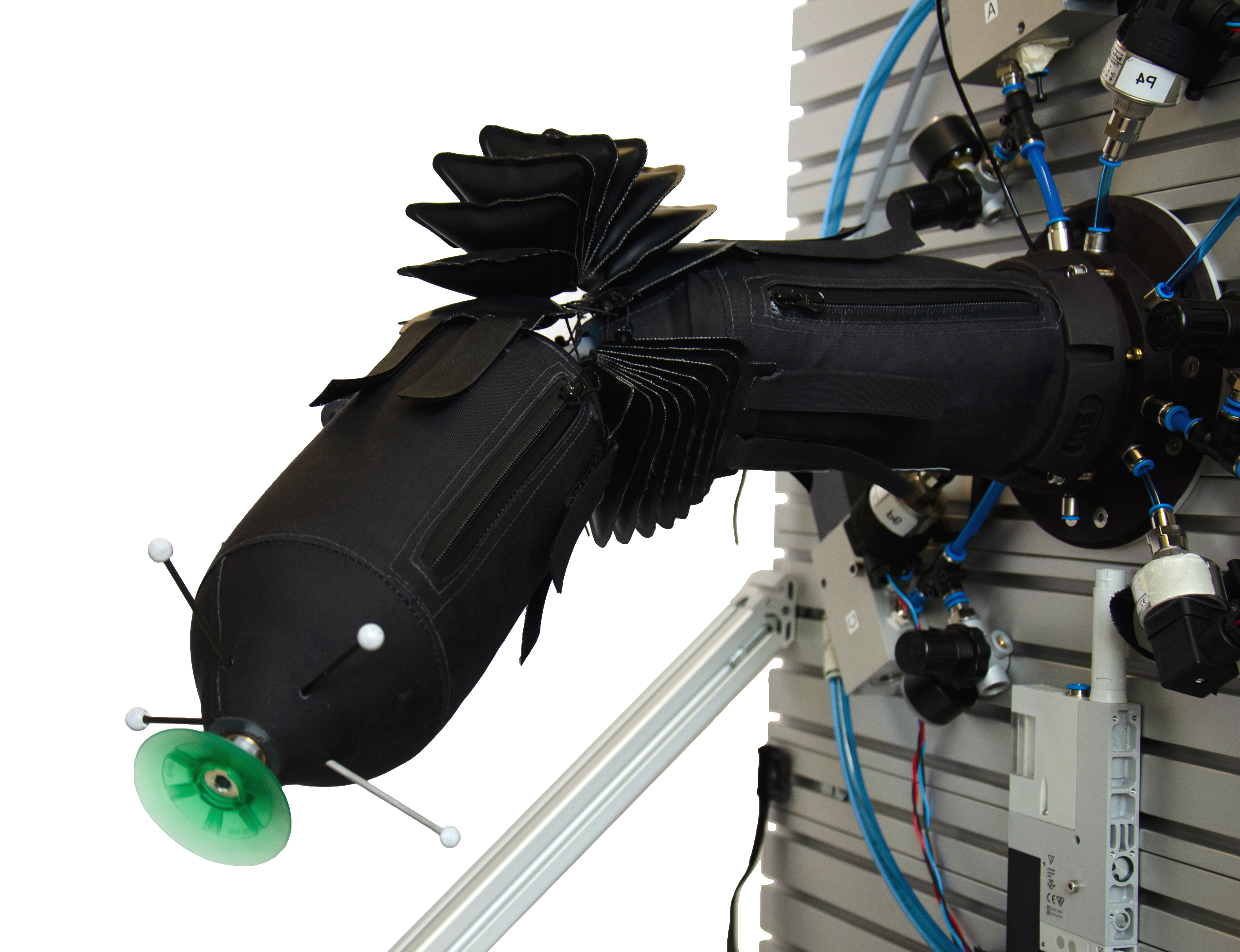}
	\vspace{\spacetocaption}
	\caption{The soft robotic arm used for the experimental evaluation. The arm is actuated by three symmetrically arranged bellow-type actuators. The arm includes a suction cup as an end effector and is mounted in a horizontal configuration, which is used throughout this work.  }
	\label{Fig:SoftRoboticArm}
	\vspace{\spacetofig}
\end{figure}

In terms of pick-and-place applications with soft robotic systems,  a grasp-and-place operation using a planar manipulator consisting of a series of fluidic arm segments is outlined in \cite{katzschmann_graspAndPlace}. A pick-and-place application is described in \cite{hyattPP_space} using a soft robotic arm consisting of a series of antagonistically actuated single degree of freedom inflatable segments. In \cite{rusILCsoft_grasping}, a grabbing task based on ILC is implemented on a  soft spatial fluidic elastomer manipulator. Continuum robots for object manipulation are documented in \cite{festoBionicHandling} and \cite{polygerSoftPolyLimb}.    

In this work, we present the realization of a pick-and-place application using the antagonistically actuated spherical soft robotic arm depicted in Fig. \ref{Fig:SoftRoboticArm}. Two rotational and one translational or stiffness degree of freedom can be simultaneously controlled in a single joint based on the continuous structure of the soft joint. 

First, a control allocation based on pressure differences and the antagonistic actuation principle of the system is used to decouple the plant dynamics, simplifying the modeling and control. An LPV model is identified, capturing the influence of the load mass and a parameter which is related to the stiffness of the joint. A gain-scheduled feedback controller is designed, which relies on the parameter-varying dynamics and which ensures robustness over a large range of the parameters considered. This property is validated by manipulating an object with similar weight to the movable link of the soft robot. As pick-and-place applications are often repetitive in nature, an ILC scheme is proposed to enable accurate tracking of aggressive maneuvers. The simultaneous control of elongation and angle is shown, leveraging the compliance of the soft joint. This feature is required when picking an object with the suction cup. Different to the work presented in \cite{katzschmann_graspAndPlace}, \cite{hyattPP_space}, and \cite{rusILCsoft_grasping}, we assume the locations and the trajectory to be fixed during the pick-and-place application, restricting the pick-and-place task to repetitive applications. However, the learning control methodology presented can be applied to an arbitrary trajectory and allows us to realize a fast and reliable pick-and-place application. 
 
The remainder of this paper is organized as follows: The soft robotic platform is briefly described in Section \ref{sec:SoftRoboticArm}. A control allocation based on pressure differences to decouple the plant dynamics is introduced in Section \ref{sec:Decoupling}. The identification of the LPV model is presented in Section \ref{sec:Modeling_SysID} and the design of a gain-scheduled feedback controller and an ILC scheme is outlined in Section \ref{sec:Control}. The realization of the pick-and-place application is discussed in Section \ref{sec:Application} and a conclusion is drawn in Section \ref{sec:Conclusion}.

\section{Soft Robotic Arm}\label{sec:SoftRoboticArm}
In this section, the spherical soft robotic arm used for the pick-and-place application  is described. 

\begin{figure}
	\centering
	\begin{minipage}[t]{0.54\columnwidth}
		\begingroup
		\tikzset{every picture/.style={scale=0.92}}

\usetikzlibrary{shadings,intersections}
\newcommand{\arr}{-{Latex[length=1mm, width=1mm]}}
\newcommand{\arrd}{{Latex[length=1mm, width=1mm]}-{Latex[length=1mm, width=1mm]}}

\begin{tikzpicture}[scale = 1]

\begin{scope}[rotate around={90:(0,0)}]

\draw[very thick] (0,0) -- (0,2);
\draw[very thick] (0,2) -- (0.2016,3.5072); 
\draw[dashed] (0,2.05) -- (0,4.52);
\draw[dashed] (0.005,2.01) -- (-0.4,3.7);
\draw[very thick, \arr] (0.2184, 3.6328) -- (0.2940, 4.1980); 
\draw[very thick,\arr] (-0.4,3.7) arc [radius=1.5, start angle=90, end angle= 68];        
\draw [very thick,\arr] (0.0,4.40) arc [radius=1.25, start angle=132, end angle= 170]; 
\draw[\arr] (0,0) -- (1,0);     
\draw[\arr] (0,0) -- (0.6,0.6); 
\node at (0.6,0.8) {$\vec{e}_y$};
\node at (1.17,0.11) {$\vec{e}_x$};
\node at (-0.2,3.48) {$\beta$};
\node at (-0.45,4.1) {$\alpha$};
\node at (0.5,4.0) {$R$};
\draw [fill] (0,2) circle [radius=0.06];
\draw [gray] (0.21,3.57) circle [radius=0.06]; 
\draw [] (0.3024, 4.2608) circle [radius=0.06];

\coordinate (O) at (0,2);

\draw[ultra thin] (0,2) to [edge label = $$] (-1.57,3.37);
\draw[ultra thin] (0,2) -- (1.57,3.37);

\draw[] (-1.57,3.46) arc [start angle = 140, end angle = 40,
  x radius = 20.4mm, y radius = 26mm];
\draw[densely dashed] (-1.559,3.46) arc [start angle = 170, end angle = 10,
  x radius = 15.8mm, y radius = 3.6mm];
\draw[] (-1.489,3.52) arc [start angle=-200, end angle = 20,
  x radius = 15.8mm, y radius = 3.15mm];

\end{scope}

\end{tikzpicture}
		\endgroup
	\end{minipage}
	\begin{minipage}[t]{0.44\columnwidth}
		\begingroup
		\tikzset{every picture/.style={scale=0.73}}
		\usetikzlibrary{shadings,intersections}
\newcommand{\arr}{-{Latex[length=1mm, width=1mm]}}
\newcommand{\arrd}{{Latex[length=1mm, width=1mm]}-{Latex[length=1mm, width=1mm]}}

\begin{tikzpicture}[scale = 1]

\begin{scope}[rotate around={180:(0,0)}]

\draw[\arr] (5.5,2) -- (6.5,2);
\draw[\arr] (4,0.5) -- (4,-0.5);
\draw[ultra thin] (5.5,2) -- (4,2) -- (4,0.5);
\node at (3.7,-0.6) {$\vec{e}_x$};
\node at (6.6,2.15) {$\vec{e}_y$};

\begin{scope}[rotate around = {30:(4,2)}]
\draw[ultra thin] (4,2) -- (6,2);
\end{scope}
\begin{scope}[rotate around = {150:(4,2)}]
\draw[ultra thin] (4,2) -- (6,2);
\end{scope}
\draw [\arrd,ultra thin,domain=30:150] plot ({4+1.7*cos(\x)}, {2+1.7*sin(\x)});
\node at (4,3.52) {\tiny $120^{\circ}$};
\draw [\arrd,ultra thin,domain=150:270] plot ({4+1.7*cos(\x)}, {2+1.7*sin(\x)});
\node at (2.79,1.29) {\tiny $120^{\circ}$};
\draw [\arrd,ultra thin,domain=270:390] plot ({4+1.7*cos(\x)}, {2+1.7*sin(\x)});
\node at (5.20,1.29) {\tiny $120^{\circ}$};

\draw (3.5,1.1) -- (3.8,1.88) -- (4.2,1.88) -- (4.5,1.1);
\draw (3.6,0.9) -- (4.4,0.9);
\draw[] (4.5,1.1) to [out=-70,in=0] (4.4,0.9);
\draw[] (3.5,1.1) to [out=-110,in=180] (3.6,0.9);
\node at (3.78,1.13) {A};

\begin{scope}[rotate around = {120:(4,2)}]
\draw (3.5,1.1) -- (3.8,1.88) -- (4.2,1.88) -- (4.5,1.1);
\draw (3.6,0.9) -- (4.4,0.9);
\draw[] (4.5,1.1) to [out=-70,in=0] (4.4,0.9);
\draw[] (3.5,1.1) to [out=-110,in=180] (3.6,0.9);
\end{scope}
\node at (4.88,2.23) {B};

\begin{scope}[rotate around = {240:(4,2)}]
\draw (3.5,1.1) -- (3.8,1.88) -- (4.2,1.88) -- (4.5,1.1);
\draw (3.6,0.9) -- (4.4,0.9);
\draw[] (4.5,1.1) to [out=-70,in=0] (4.4,0.9);
\draw[] (3.5,1.1) to [out=-110,in=180] (3.6,0.9);
\end{scope}
\node at (3.35,2.68) {C};

\end{scope}

\end{tikzpicture}
		\endgroup
	\end{minipage}
	\vspace{\spacetocaption}
	\caption{The left plot shows the parametrization of the robot orientation using extrinsic Euler angles $\alpha, \beta$, both describing rotations with respect to the inertial frame. The axial degree of freedom enabled by the axial compliance of the soft joint is represented by the (adjustable) radius $R$. The antagonistic actuator configuration in the respective coordinate system is shown in the right plot. The gravitational vector points in the negative $\vec{e}_x$ direction. }
	\label{fig:orientation_coFrame_def}
	\vspace{\spacetofig}
\end{figure}
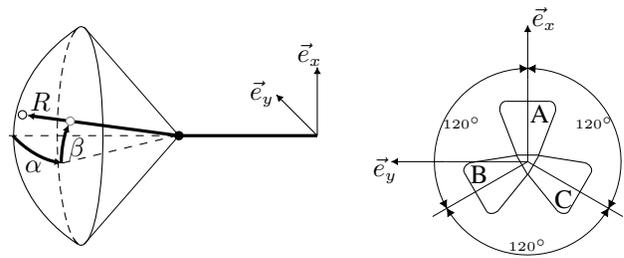

The soft robotic arm consists of six main components: a static link, a movable link, a soft joint, three bellow type actuators, a vacuum gripper as an end effector and a base plate for mounting the system (see Fig. \ref{Fig:SoftRoboticArm}). The static- and movable links are based on a double-shell design, consisting of an inflatable, airtight inner bladder and a sewn, inextensible outer shell. These result in a lightweight construction, where the mass of the movable link is given by \unit[200]{g}. Pressurization of the links provides the structural stability of the robot arm. The soft joint connects the static- with the movable link and is symmetrically enclosed by the actuator triplet, as depicted in Fig. \ref{fig:orientation_coFrame_def} (right). The design is based on previous work presented in \cite{mhofer:DesFabModCtrl_V3}, with several components and properties of the system being optimized to provide the mechanical robustness required for the realization of a pick-and-place application. This particularly applies to the actuator design. The maximum burst pressure of the actuators is increased from 2.2 to \unit[6.0]{bar} by using a material with high tensile strength (Rivertex® Riverseal® 842) and a more sophisticated manufacturing process based on high-frequency welding. Furthermore, the joint system of the soft robotic arm is optimized. Different to the previous system, which relied on a rigid ball-and-socket joint, a soft joint made from silicone rubber (Wacker Elastosil\textregistered $\,$ M 4641) is used, enabling large deformations. The soft joint has several important features. On the one hand, it allows for an extension of the angular range from 45$^\circ$ to $75^\circ$. Furthermore, it introduces a controllable, translational degree of freedom (see Fig. \ref{fig:orientation_coFrame_def}, left) and allows us to route the tubing for the air supply internally. The additional degree of freedom is a key feature when picking an object in a pick-and-place application, as it enables axially approaching the object with the suction cup.

The combination of the soft joint and the antagonistic configuration of the three actuators allows for the control of two rotational degrees of freedom and the independent control of either the overall joint stiffness or the axial elongation in a single joint. Notice that an increase in the overall joint stiffness inevitably induces an increase in the axial elongation and vice versa. That is, the joint stiffness and the axial elongation cannot be controlled independently, as the number of control inputs is restricted to three. 

The orientation $(\alpha, \beta)$ of the robot arm is retrieved from a Vicon motion capture system with sub-millimeter accuracy. Proportional valves are deployed to control the pressure in the actuators. A vacuum ejector is used to generate the vacuum as required for the operation of the end effector.      

\section{Control Allocation}\label{sec:Decoupling} 
Directly applying the pressures $p_A, p_B,$ and $p_C$ to the system would simultaneously affect the orientation and the joint stiffness of the robot arm. Therefore, we introduce a control allocation which allows us to represent a point $(p_A, p_B, p_C)$ by two variables related to the angular deflection of the arm and a variable connected to the joint stiffness. To this end, we first generalize the concept of pressure difference to the case of three pressures. Based on this, a mapping is derived which decouples the two angular degrees of freedom, $\alpha, \beta$, allowing us to simplify the modeling and control approach. Finally, the proposed mapping is experimentally validated.       

\subsection{Pressure Difference}
The motivation to rely on pressure differences becomes clear when considering a system with a single degree of freedom with antagonistic actuation (see \cite{gillespie:simPosStiffCtrl} and \cite{mhofer:DesModCtrl_V2} for examples). Due to the opposing directions of the forces applied by each actuator, the pressure difference between the two actuators is directly coupled to the angular deflection.

For the case of three independent actuators A, B, and C and corresponding pressures $p_A, p_B$, and $p_C$, we define the pressure differences as 
\begin{align}
	\Delta p_{AB} \coloneqq p_A - p_B, \label{eq:pDiff_AB} \qquad 
	\Delta p_{BC} \coloneqq p_B - p_C.
\end{align}
Note that this definition is arbitrary as long as each pressure occurs at least once in the equations. The difficulty lies in the determination of the inverse mapping, i.e. the relation from $\Delta p_{AB}, \Delta p_{BC}$ to $p_A, p_B, p_C$. The inverse mapping is straightforward in the case of two pressures and presented in \cite{mhofer:DesModCtrl_V2}. In the following, we present a way to generalize the concept to the case of three actuators. We introduce the linear system of equations 
\begin{align}
	\begin{bmatrix}
		1 & -1 & 0 \\ 0 & 1 & -1
	\end{bmatrix}
	\begin{bmatrix}
		p_A \\ p_B \\ p_C 
	\end{bmatrix} = 
	\begin{bmatrix}
	\Delta p_{AB} \\ \Delta p_{BC} 
	\end{bmatrix} \label{eq:LSE}
\end{align}
constrained by 
\begin{align}
	\bar{p} = \min(p_A, p_B, p_C), \qquad p_i \ge \bar{p}, \,\, i=A,B,C. \label{eq:constraints_LSE}
\end{align}
One can show that the solution of \eqref{eq:LSE}-\eqref{eq:constraints_LSE} is uniquely determined by
\begin{align}
	p_A &= \max\{ \bar{p}, \bar{p} + \Delta p_{AB}, \bar{p} + \Delta p_{AB} + \Delta p_{BC} \} \label{eq:delta2abs_A} \\
	p_B &= \max\{ \bar{p}, \bar{p} + \Delta p_{BC}, \bar{p} -\Delta p_{AB} \} \\
	p_C &= \max\{ \bar{p}, \bar{p} - \Delta p_{BC}, \bar{p} -\Delta p_{AB} -\Delta p_{BC} \}. \label{eq:delta2abs_C}
\end{align} 
The parameter $\bar{p}$ corresponds to the lower pressure level in all three actuators, similarly used in \cite{gillespie:simPosStiffCtrl} and \cite{mhofer:DesFabModCtrl_V3} to adjust the stiffness of the joint for a single or two degrees of freedom system, respectively. 
The inverse mapping defined by \eqref{eq:delta2abs_A}-\eqref{eq:delta2abs_C} ensures that $p_i \ge \bar{p} \,\,, i = A, B, C$ at all times. Notice that at least one of these inequality constraints is always active. Furthermore, note that the differences $p_A - p_B$ and $p_B - p_C$  result in $\Delta p_{AB}$ and $\Delta p_{BC}$ as desired, when applying \eqref{eq:delta2abs_A}-\eqref{eq:delta2abs_C}.

\subsection{Decoupling}
The mapping introduced in the previous section allows us to uniquely represent three absolute pressures $p_A, p_B, p_C$ using two variables $\Delta p_{AB}, \Delta p_{BC}$ which are related to an excitation in the $\alpha$-$\beta$-plane and one variable $\bar{p}$ which is related to stiffness. The pressure differences $\Delta p_{AB}, \Delta p_{BC}$ have associated directions of action resulting from the configuration of the actuators. The associated directions of action of $\Delta p_{AB}, \Delta p_{BC}$ can be explicitly computed by introducing a kinematic model of the soft robotic arm. A kinematic relationship between the pressures and their effect on $\alpha, \beta$, incorporating the actuator configuration (see Fig. \ref{fig:orientation_coFrame_def}), assuming small values for $\alpha$ and $\beta$, is introduced in \cite{mhofer:DesFabModCtrl_V3}, namely, 
\begin{align}
	\alpha \, &\propto \, \frac{\sqrt{3}}{2} p_{B} -\frac{\sqrt{3}}{2} p_C \label{eq:propto_1}\\
	\beta \, &\propto \, -p_A + \frac{1}{2} p_B + \frac{1}{2} p_C.\label{eq:propto_2}
\end{align}
Combining the definitions in \eqref{eq:pDiff_AB} with \eqref{eq:propto_1}-\eqref{eq:propto_2} allows us to define the decoupled pressure differences $\Delta p_\alpha$ and $\Delta p_\beta$, which are aligned with the orientations $\alpha$ and $\beta$, namely, 
\begin{align}
\begin{bmatrix} \Delta p_\alpha \\ \Delta p_\beta \end{bmatrix}
	\coloneqq 	\begin{bmatrix}
		0 & \sqrt{3}/2 \\ -1 & -1/2 
	\end{bmatrix}
	\begin{bmatrix}
	\Delta p_{AB} \\ \Delta p_{BC} 
	\end{bmatrix}. \label{eq:lin_transf}
\end{align}
Note that the linear transformation is invertible.  

\subsection{Composition and Experimental Validation}
\begin{figure}
	\begin{minipage}[t]{0.49\columnwidth}
		\centering
		\includegraphics[scale=0.32]{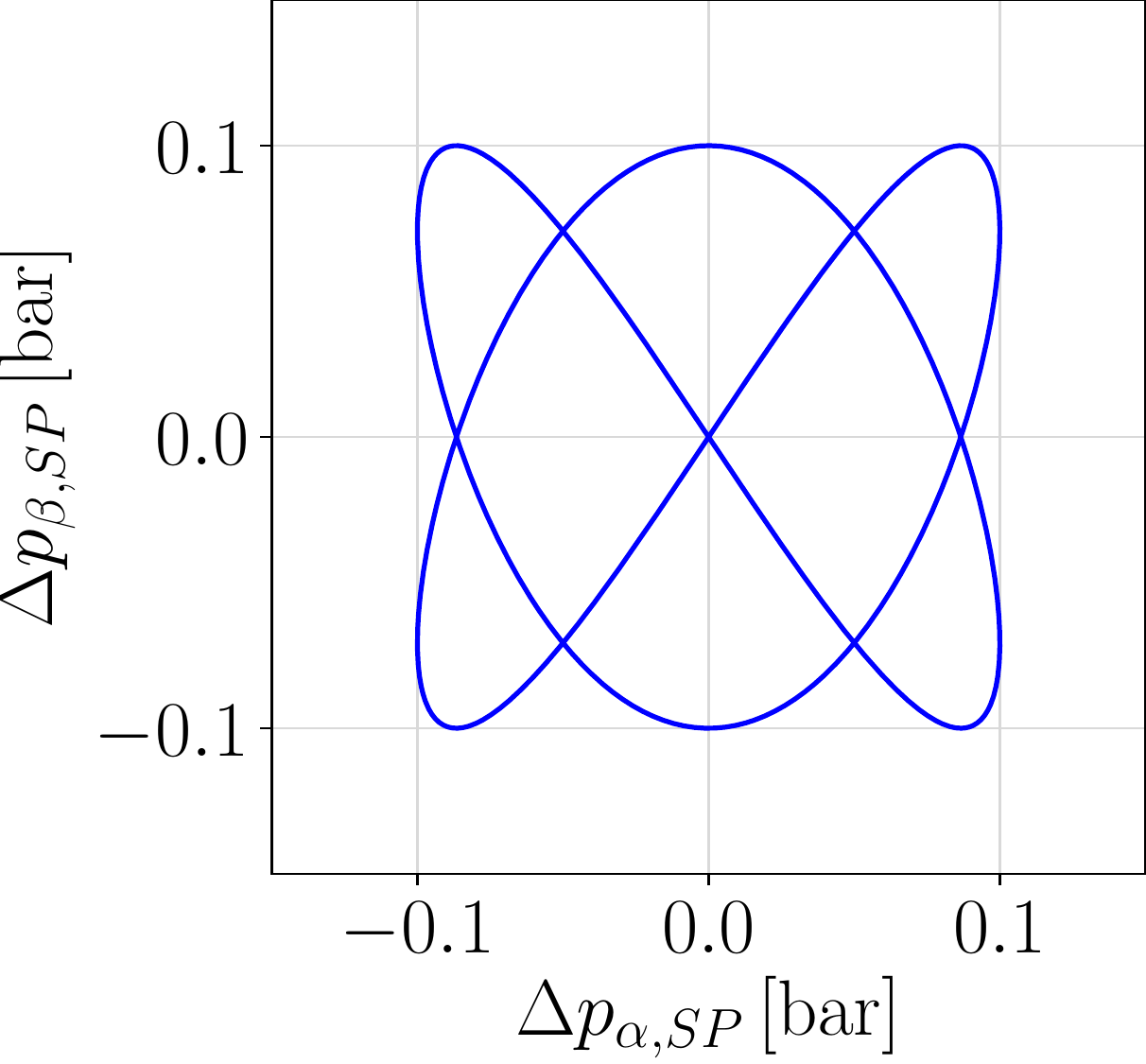}
	\end{minipage}
	\begin{minipage}[t]{0.49\columnwidth}
		\centering
		\includegraphics[scale=0.32]{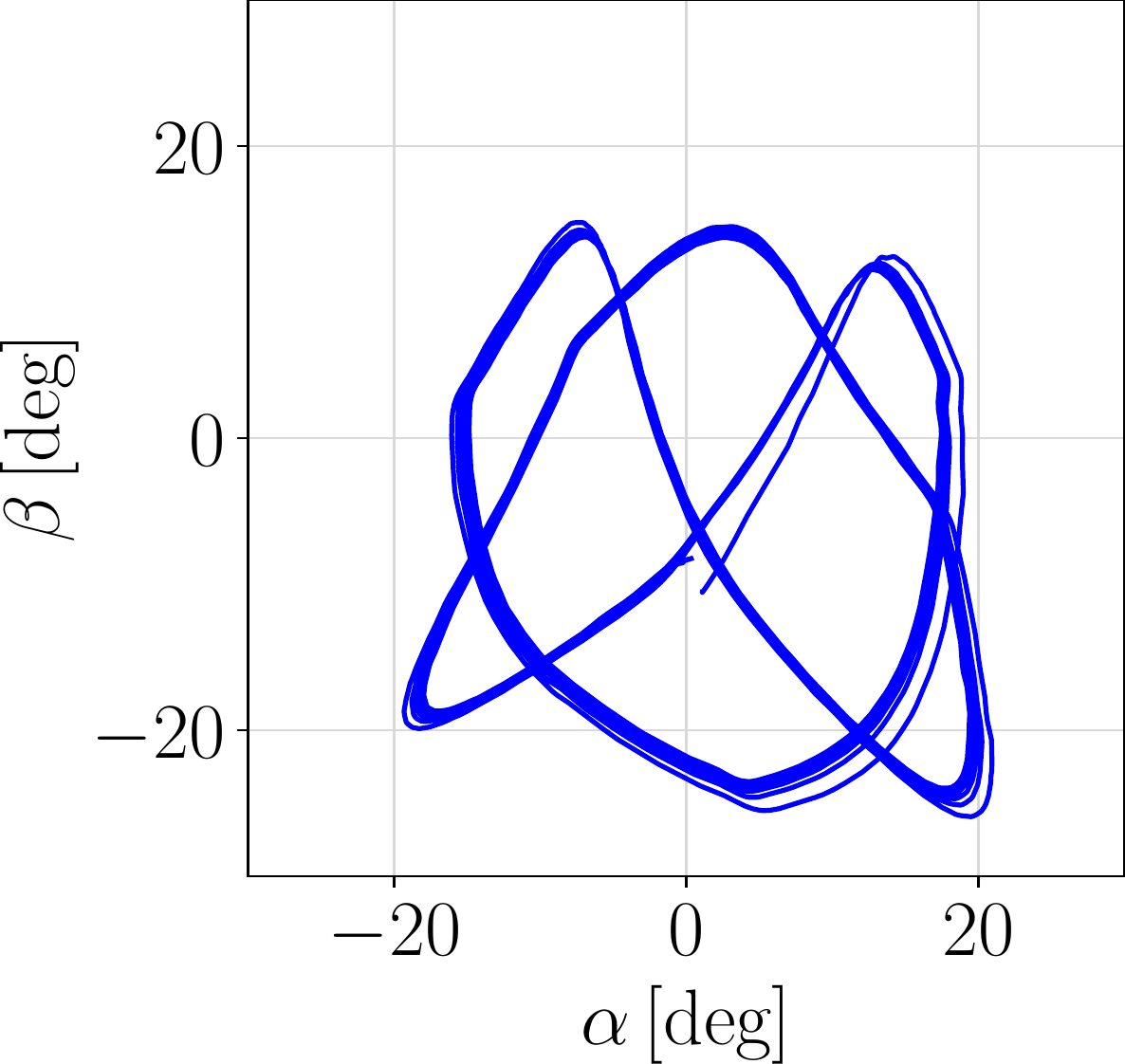}
	\end{minipage}
	\vspace{\spacetocaption}
	\caption{The figures show an experimental validation of the decoupling property of the Delta Representation. The left plot shows the applied reference trajectory in the Delta Representation space ($\bar{p}$ is set to 1.05 bar), whereas the right plot shows the resulting trajectory in the angle space. The frequency ratio and phase shift is fixed such that a so-called Lissajous curve results. The pressure trajectories are applied over multiple periods showing the high repeatability of the system within one experiment. Imperfections in the similarity property are caused by viscoelastic material behavior.   }
	\label{fig:lissaj_space_valExp}
	\vspace{\spacetofig}
\end{figure}

The composition of \eqref{eq:delta2abs_A}-\eqref{eq:delta2abs_C} and the inverse of \eqref{eq:lin_transf} allows us to define a bijective mapping, denoted as 
\begin{align}
	(p_A, p_B, p_C)  = \xi (\bar{p}, \Delta p_\alpha, \Delta p_\beta). \label{eq:xi_deltaRep}
\end{align}
A point $(\bar{p}, \Delta p_\alpha, \Delta p_\beta)$ is referred as the Delta Representation of a corresponding point in the absolute pressure space $(p_A, p_B, p_C)$. A change in $\Delta p_\alpha$ solely affects the orientation $\alpha$ and analogously for $\Delta p_\beta$ and $\beta$, independent of $\bar{p}$. For the purpose of validation, sinusoidal trajectories with respect to the virtual control inputs $\Delta p_\alpha$ and $\Delta p_\beta$ are applied. The corresponding pressure setpoint trajectories (for a fixed value of $\bar{p}$) are computed offline using \eqref{eq:xi_deltaRep} and tracked by three independent pressure controllers. An example for the trajectory in the Delta Representation space and the resulting trajectory in the angle space ($\alpha, \beta$) is shown in Fig. \ref{fig:lissaj_space_valExp}. Note that no angle control is applied. The use of the Delta Representation induces a similarity property between the angle and pressure space, allowing us to consider the dynamics as decoupled. The Delta Representation has several desirable properties which are discussed in more detail in the following sections. For a detailed mathematical analysis and further experimental validation of the Delta Representation, the reader is referred to the Online Appendix \cite{zughaibi_onlineAppx}.  
\graphicspath{{img/}}


\section{Modeling}
\label{sec:Modeling_SysID}
In this section, the modeling and parameter identification of the soft robotic arm are presented. An LPV model is developed, which captures the fundamental behavior of the system. The underlying requirement is to be sufficiently descriptive for controller synthesis, assuming inaccuracies and model uncertainties can be compensated by feedback control and a learning control strategy (see Sec. \ref{sec:Control}). 

Because of the lightweight structure of the robotic arm, the load mass, $m$, has a significant influence on the dynamics of the system. This becomes particularly important given the objective of carrying a significant load mass during a pick-and-place application. Therefore, the dependency of the load mass is explicitly considered in the modeling process. Similarly, the parameter for the lower pressure bound $\bar{p}$ has a significant influence on the system dynamics due to its strong connection to the joint stiffness. As a consequence of defining the input in the Delta Representation space, the model structure is assumed to be diagonal, namely, 
\begin{align}
	\begin{bmatrix}
		\alpha (s) \\ \beta(s)
	\end{bmatrix} = \begin{bmatrix}
		G_\alpha(s, \bar{p}, m) & 0 \\ 0 & G_\beta(s, \bar{p}, m)
	\end{bmatrix} \begin{bmatrix}
		\Delta p_\alpha (s) \\ 
		\Delta p_\beta (s)
	\end{bmatrix}, \label{eq:model_strucutre_dec}
\end{align}
representing an LPV system with respect to the quasi statically assumed parameters $\bar{p}$ and $m$, where $s$ denotes the Laplace variable. The modeling process can be divided into two steps. First, a gray box model is derived, allowing us to incorporate first principles knowledge. Secondly, the parameters are estimated using a frequency domain identification procedure.  

\subsection{Gray Box Model}
We consider the system as two independent (linearized) pendulums. Given the robot configuration as depicted in Fig. \ref{Fig:SoftRoboticArm}, the linearized dynamics are given by,
\begin{align}
	\biggl(m R_0^2 + M \frac{R_0^2}{4}\biggr) \ddot{\alpha} + d_\alpha \dot{\alpha} + k_\alpha \alpha &= \tau_\alpha, \label{eq:eom_horiz_phi} 
\end{align}
and analogously for $\beta$. The parameter $R_0$ represents the radius of the movable link from the pivot point to the load mass, when the soft joint is not extended. The second inertia term in \eqref{eq:eom_horiz_phi} accounts for the inertia of the movable link, where a distance of $R_0/2$ is assumed from the pivot point to the center of mass of the movable link without any load mass attached. The mass of the movable link is denoted by $M$.  
The stiffness and damping parameters $k_i, d_i$ are unknown and estimated from experimental data as discussed in the subsequent paragraph. The mass-related parameters are measured with a weighing scale. The driving torque $\tau_i$ is assumed to be a linear dynamic system in the decoupled pressure difference, namely, 
\begin{align}
	\tau_i(s) = H_i(s, \bar{p}) \Delta p_i(s)  \,\,, i = \alpha, \beta, \label{eq:driving_torque_tf_gen}
\end{align}
where the order and the parameters of the transfer function are determined from identification experiments. It is assumed that the parameters of $H_i$, as well as the stiffness and damping coefficients $k_i$, $d_i$, solely depend on $\bar{p}$ and are independent of $m$. As a consequence, the parameter estimation experiments can be conducted for a fixed load mass $m$, as its influence can be extrapolated from the first principles model in \eqref{eq:eom_horiz_phi}. The system identification experiments only address variations with respect to $\bar{p}$, which significantly reduces the number of experiments required.

\subsection{Parameter Estimation}
\begin{figure}
	\centering
	\includegraphics[scale=\pythonfigwidth]{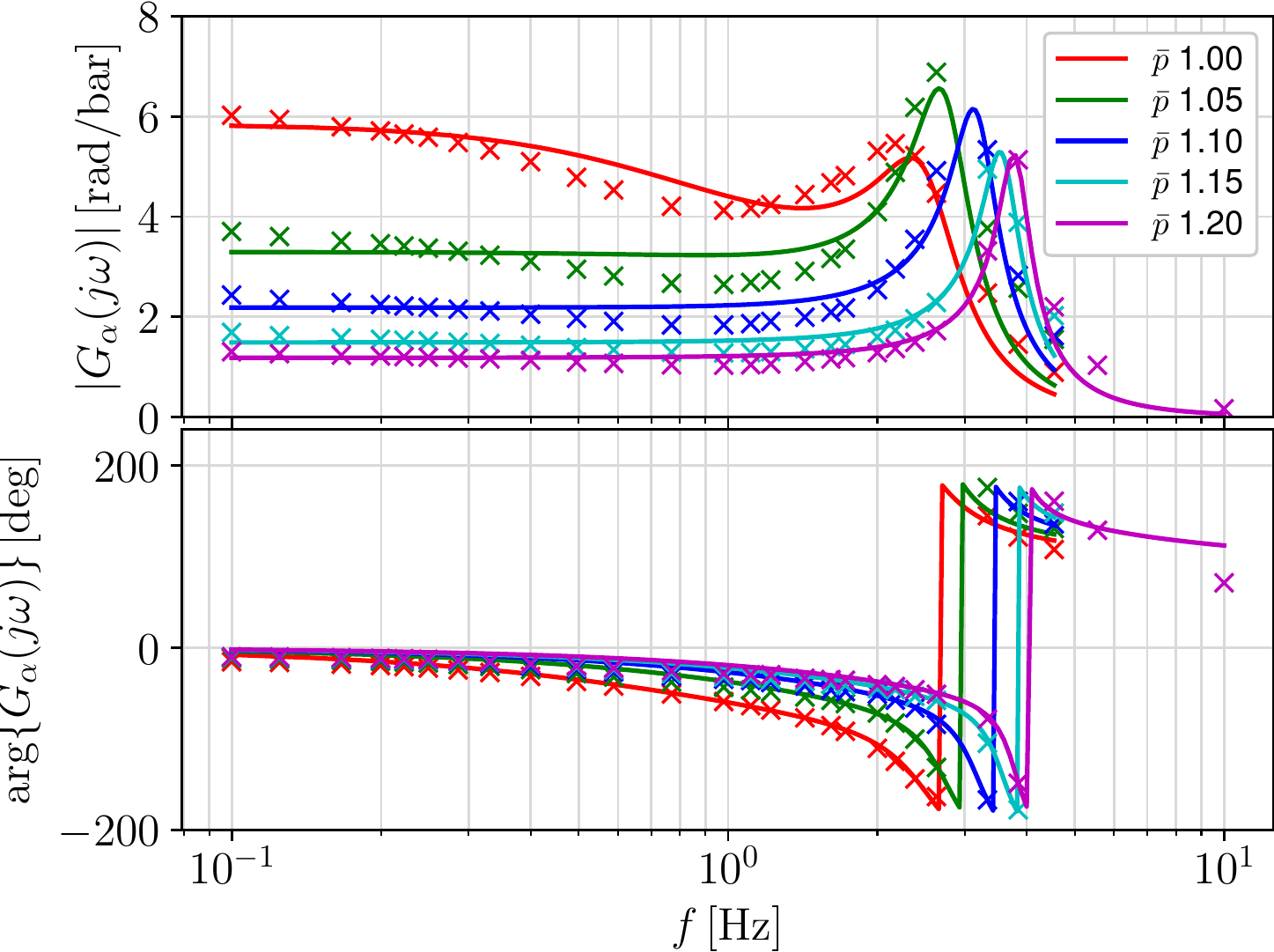}
	\vspace{\spacetocaption}
	\caption{Frequency response data obtained in the identification experiments for $G_\alpha$ for different values of $\bar{p}$ and their corresponding fits. Note the right shift of the resonance frequency for increasing $\bar{p}$, as expected for increasing stiffness. Similar results are obtained for $G_\beta$. The estimates close to the resonance frequency are characterized by higher uncertainty due to couplings arising in the pressure dynamics and the excitation of flexible modes of the robotic links.}
	\label{fig:bode_data_fit_alpha}
	\vspace{-6pt}
\end{figure}
\begin{figure}
	\centering
	\includegraphics[scale=\pythonfigwidth]{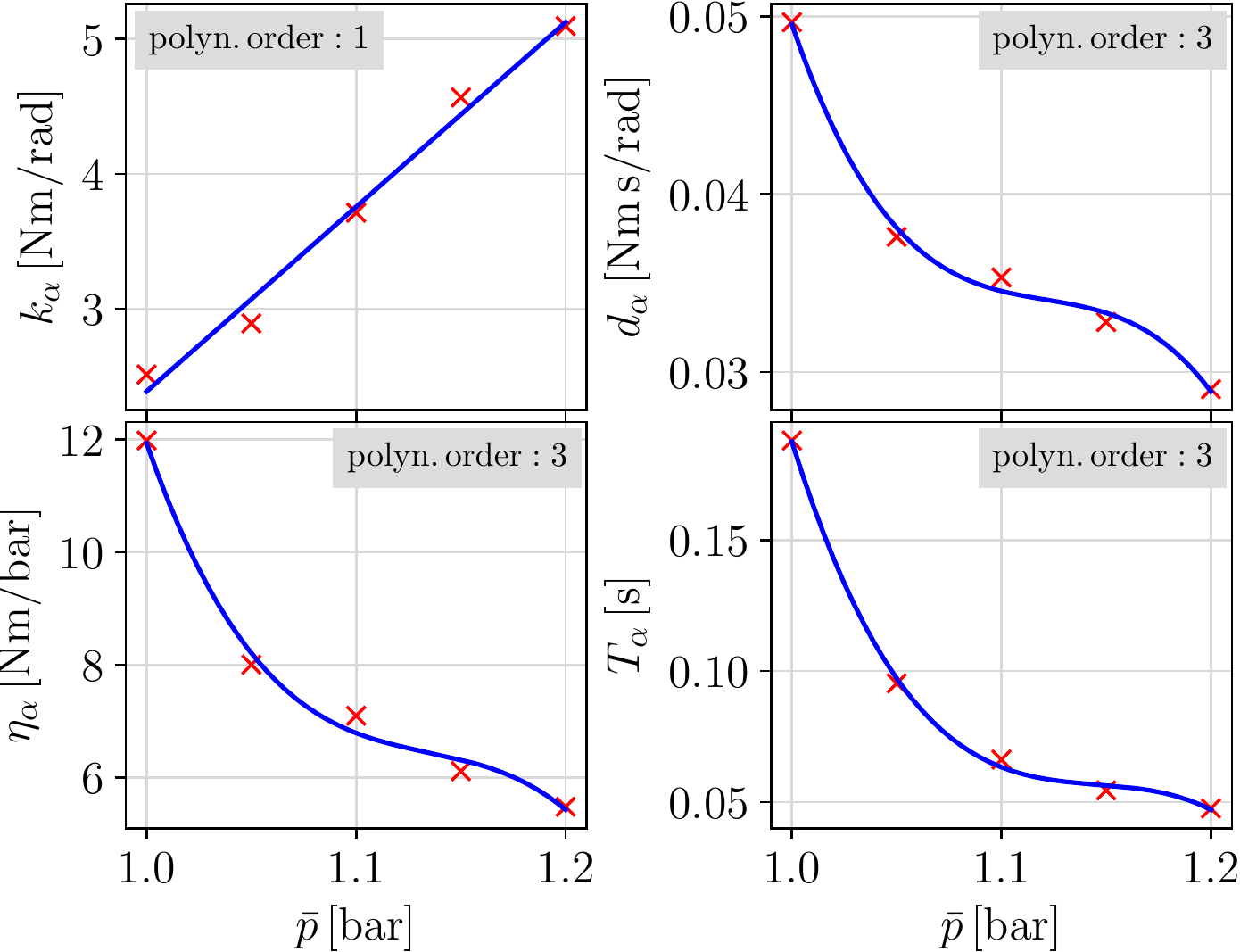}
	\vspace{\spacetocaption}
	\caption{Stiffness, damping, gain, and time constant parameters of the estimated transfer functions with respect to $G_\alpha$ as a function of $\bar{p}$. Polynomial fitting allows us to obtain a fully parametric model in terms of $\bar{p}$. A strong linear coherence between $\bar{p}$ and the stiffness coefficient $k_\alpha$ can be observed. }
	\label{fig:prm_fit_11}
	\vspace{\spacetofig}
\end{figure}
In order to estimate the parameters of the differential equations \eqref{eq:eom_horiz_phi} and \eqref{eq:driving_torque_tf_gen}, system identification experiments are conducted. The choice of the excitation signal is simplified as the input of the plant is defined in the Delta Representation space. This allows the excitation signal to attain both positive and negative values. The system is excited with a series of sinusoidal inputs providing a high signal to noise ratio as the entire signal energy is concentrated at a single frequency. For each frequency, the input signal is repeated over ten periods. The first four periods are discarded to minimize the influence of transients. The remaining periods are averaged in order to reduce the variance of the estimate. The magnitude and phase of the output response are estimated using a sinusoidal correlation method, as described in \cite{ljung_sysID}. The experiments are conducted for five different values of $\bar{p}$ and no load mass attached ($m = 0$). The Bode plot of the identified frequency response data and the corresponding fits are shown in Fig. \ref{fig:bode_data_fit_alpha}. Complex curve fitting as described in \cite{ljung_sysID} is used to fit the parameters of the transfer functions. The best fit for $G_\alpha$ and $G_\beta$ is obtained for an order of three. As the order of the equation of motion (\ref{eq:eom_horiz_phi}) is two, the order of the driving torque transfer function (\ref{eq:driving_torque_tf_gen}) is consequently given as one. The overall transfer function from the decoupled pressure difference to the angle is given by, 
\begin{align}
	G_i(s, \bar{p}, m) = \frac{\eta_i}{T_i s + 1} \frac{1}{(m + M / 4)R_0^2 s^2  + d_i s +  k_i }, \label{eq:parameteric_model_horiz}
\end{align} 
for $i = \alpha, \beta$. In order to obtain a fully parametric model, polynomials are fitted for the parameters $\eta_i = \eta_i(\bar{p}), T_i = T_i(\bar{p}), d_i = d_i(\bar{p}), k_i = k_i(\bar{p})$ for $i = \alpha, \beta$ as a function of $\bar{p}$. The corresponding fits are shown in Fig. \ref{fig:prm_fit_11}. The result is a fully parametric LPV model with respect to the parameters $\bar{p}$ and $m$.

\section{Control}
\label{sec:Control}
\begin{figure*}
	\centering
	\includegraphics[clip, trim=4.6cm 14.95cm 4.55cm 2.5cm, width=0.9\textwidth]{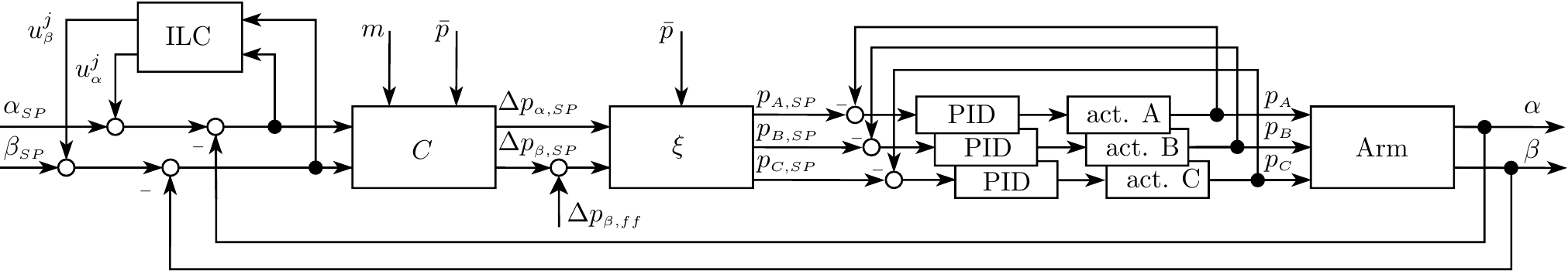}
	\caption{Block diagram showing the cascaded control architecture of the system. The output of the gain-scheduled feedback controller ($C$), running at \unit[50]{Hz}, is in the Delta Representation space and transferred to absolute pressures using \eqref{eq:xi_deltaRep}. The actuator pressures $p_A, p_B, p_C$ are controlled by three independent PID controllers, running at \unit[1]{kHz}. The control architecture is augmented with a serial iterative learning control scheme (ILC) to compensate for repetitive disturbances.   }
	\label{fig:blockdiagram}
	\vspace{\spacetofig}
\end{figure*}
In this section, the control architecture of the system is discussed. First, the design of a gain-scheduled feedback controller is presented, based on the LPV model introduced in the previous section. In the second part, an ILC strategy is presented which can compensate for repetitive errors enabling accurate tracking of aggressive trajectories. These repetitive errors arise from nonlinearities in the system dynamics, such as unmodeled actuator couplings, large deviations from the linearization point, and repetitive viscoelastic phenomenons. The tracking performance of both the feedback controller and the ILC is evaluated with and without a load mass attached to the robot arm to validate its generalization to different load masses. 

A cascaded control architecture is employed as shown in Fig. \ref{fig:blockdiagram}. The outputs of the angle controller are the setpoints $\Delta p_{\alpha,SP}$, $\Delta p_{\beta,SP}$. Consequently, the controller has no influence on the compliance of the system allowing us to consider stiffness and motion control to be independent. Accordingly, the Delta Representation addresses the issue discussed in \cite{ReductionComplianceIssue} that feedback control in general leads to a reduction in the compliance to increase the tracking performance in soft robotic systems. However, by leveraging the Delta Representation, the parameter $\bar{p}$ represents an independent degree of freedom in the control system and can be specified by the user. The setpoints in the Delta Representation are transferred to the absolute pressure space by using (\ref{eq:xi_deltaRep}). The pressure setpoints $p_{A,SP}, p_{B,SP}, p_{C,SP}$ are tracked by three independent Proportional–Integral–Derivative (PID) controllers in inner control loops.  

\begin{figure}[t!]
	\centering
	\includegraphics[scale=\pythonfigwidth]{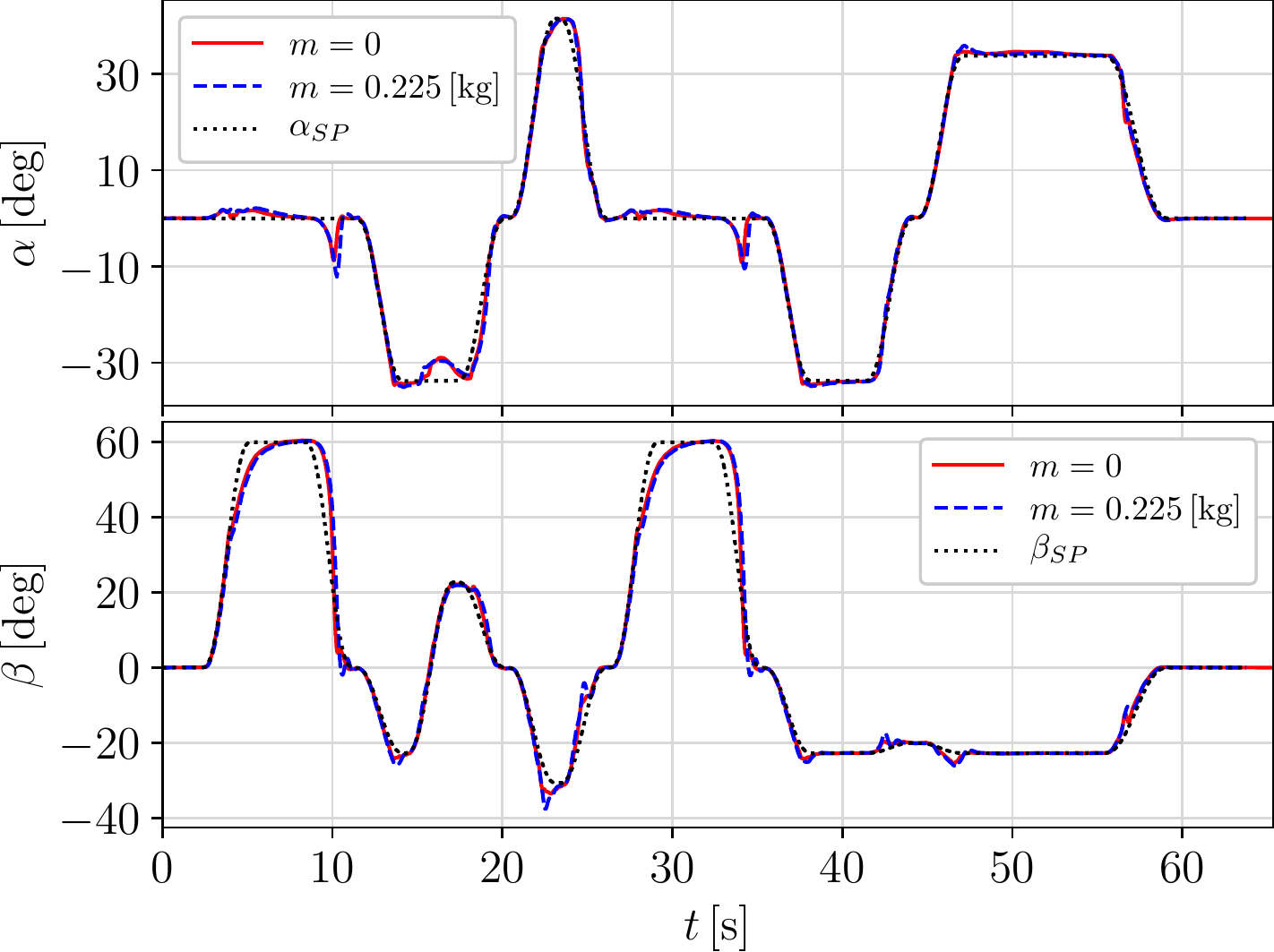}
	\vspace{\spacetocaption}
	\caption{Tracking performance of the gain-scheduled feedback controller for $\alpha$ (top plot) and $\beta$ (bottom plot) with a mass attached (blue, dashed curve) and without a mass attached to the robot arm (red, solid curve). The tracking performance is very similar irrespective of the mass attached. The controller allows for operation at high angular deflections and ensures steady state accuracy. Notice the existence of couplings between $\alpha, \beta$ for large setpoint changes and orientations far from the origin.}
	\label{fig:LPV_ref_track}
	\vspace{\spacetofig}
\end{figure}

\subsection{Feedback Control}
The feedback control structure is based on the decoupled model presented in (\ref{eq:model_strucutre_dec}). The following feedback controller is proposed, for each degree of freedom, 
\begin{align*}
	C_i(s,\bar{p}, m) = \kappa_i \frac{(m + M / 4)R_0^2 s^2  + d_i s +  k_i }{s}, i = \alpha, \beta,  
\end{align*}
which represents a linear gain-scheduled controller, parametrized by $m$ and $\bar{p}$. Assuming that $m$ and $\bar{p}$ are known, the controller $C_i(s,\bar{p}, m)$ asymptotically stabilizes the closed loop system with infinite gain margin for $\kappa_i > 0$. Note that this stability guarantee only holds for a neighborhood of the origin $(\alpha, \beta) = (0, 0)$ due to the linear nature of the model. Furthermore, the guarantee is only valid assuming the parameters $m$ and $\bar{p}$ to be constant or changing sufficiently slowly with respect to time \cite[p. 48]{briatLPVtimeDelaySys}.
A feedforward component is added, which compensates for gravitational effects acting on $\beta$ and not covered in \eqref{eq:eom_horiz_phi}, namely, 
\begin{align}
	\Delta p_{\beta, ff} = \frac{1}{\eta_\beta} \bigl( Mg \frac{R_0}{2} + m g R_0 \bigr) \cos(\beta_{SP}),
\end{align}  
where $\eta_\beta$ represents the static pressure to torque gain from \eqref{eq:driving_torque_tf_gen} and \eqref{eq:parameteric_model_horiz}, respectively.

The tracking performance of the feedback controller with and without a load mass attached is shown in Fig. \ref{fig:LPV_ref_track}. Notice the similarity in terms of tracking performance, which implies that the effect of the load mass is well captured by the first principles model and can be compensated by the pole-zero cancellation structure of the controller. The main task of the feedback controller is to ensure asymptotic stability over the entire domain of $\bar{p}$ and $m$, to provide steady state accuracy and to reject non-repetitive disturbances. However, for aggressive maneuvers, oscillations are present and the resulting overshoot is too high in order to reliably realize a pick-and-place application. As the system is characterized by a high repeatability, this effect can be compensated by the application of an ILC scheme as discussed in the following paragraph.

\subsection{Iterative Learning Control}
\graphicspath{{img/Control/}}
\begin{figure}[h]
	\centering
	\includegraphics[scale=\pythonfigwidth]{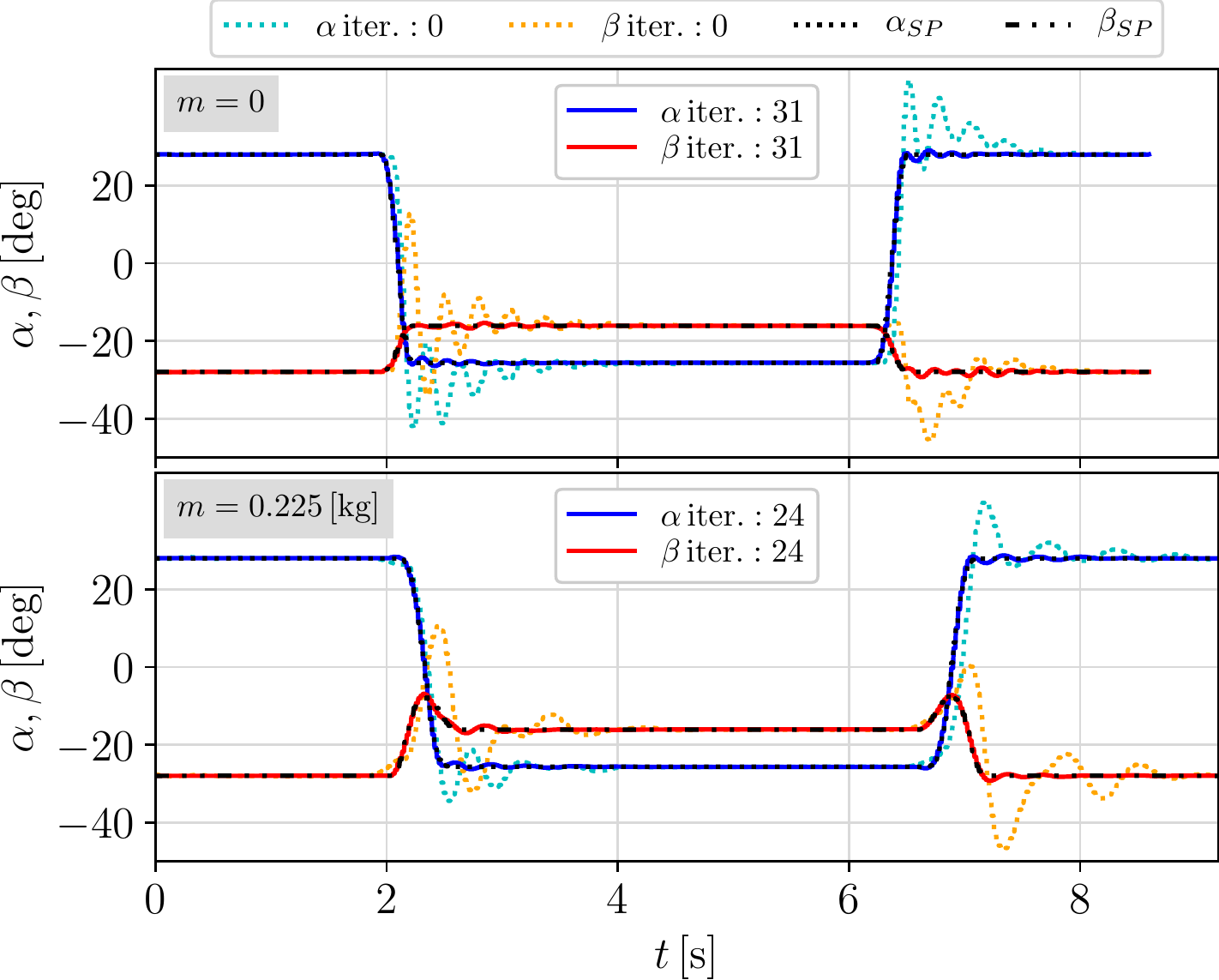}
	\vspace{\spacetocaption}
	\caption{Experimental results for aggressive trajectory tracking without load mass attached (upper plot) and with load mass attached (lower plot). For each of the two cases, the tracking performance using ILC in iteration 0 (no learning) and after the convergence of the ILC is compared. Note that the reference trajectories for the two cases are slightly different. On the one hand, the transition times for the reference change are slightly larger with the attached mass. On the other hand, the shape with respect to $\beta$ differs in order to avoid collisions with the object tray during the pick-and-place application after the object adheres to the suction cup.}
	\label{fig:ILC_noMass_Mass}
	\vspace{\spacetofig}
\end{figure}
ILC allows for accurate reference tracking in repetitive tasks \cite{bristow2006survey}, as often encountered in pick-and-place applications.
The basic idea of ILC is to adapt a feedforward/reference signal based on the error information from the previous iterations. In general, ILC can be deployed in serial or in parallel with a feedback controller. In the following, a norm-optimal iterative learning controller in a serial architecture is employed \cite{owensILCoptimization}. A serial architecture simplifies the design process of the ILC as it allows us to assume the dynamics to be parameter-independent, given the feedback controller structure proposed above. In particular, the closed loop dynamics are independent of the parameter $m$ due the pole-zero cancellation in the open-loop system. Note that the feedback controller structure cannot fully compensate the influence of $\bar{p}$, in particular the driving torque dynamics in (\ref{eq:driving_torque_tf_gen}). However, an analysis of the frequency response reveals that the closed loop dynamics are similar with respect to $\bar{p}$ over a large frequency range. Hence, the closed loop dynamics are assumed to be independent of $\bar{p}$, using an average value of $\bar{p} = \unit[1.10]{bar}$ of the considered range $\bar{p} \in [1.0, 1.2]$\hspace{0.5mm}bar to obtain a nominal model of the closed loop dynamics. Therefore, the same tuning matrices are used for a serial type ILC irrespective of the applied load mass, $m$, and the lower pressure level, $\bar{p}$. In the previous work \cite{mhofer:DesFabModCtrl_V3}, an ILC in a parallel architecture is implemented that is based on a much simpler model and does not leverage the Delta Representation. More specifically, in the parallel architecture from the previous work, the correction signals are added to the individual pressures $p_A, p_B, p_C$ and consequently impose an undesired change in the compliance of the system during position control. Moreover, the previous work does not cover the manipulation of any objects. 

Norm-optimal iterative learning control is a model-based ILC strategy. Based on a quadratic optimization problem the optimal correction signal is computed for each iteration. First, we introduce the following notation based on the lifted system representation, 
\begin{align*}
	u^j &\coloneqq \begin{bmatrix}
		u^j_\alpha(0), u^j_\beta(0),\ldots,u^j_\alpha(N-1), u^j_\beta(N-1)  
	\end{bmatrix}^T \\
	y^j &\coloneqq \begin{bmatrix}
		\alpha^j(0), \beta^j(0),\ldots,\alpha^j(N-1),\beta^j(N-1)  
	\end{bmatrix}^T  \\ 
	y_{SP} &\coloneqq \begin{bmatrix}
		\alpha_{SP}(0), \beta_{SP}(0),\ldots,\alpha_{SP}(N-1), \beta_{SP}(N-1)  
	\end{bmatrix}^T,
\end{align*}
where $j$ denotes the iteration index. The vector $u^j$ represents the reference correction signal of the serial ILC architecture. The error in iteration $j$ is defined as,
\begin{align}
	e^j \coloneqq y_{SP} - y^j.
\end{align}
The dynamics over an iteration $j$ can be written as, 
\begin{align}
	y^j = P u^j + P y_{SP},
\end{align}
where the second term can be considered as a repetitive disturbance. The lifted system matrix $P$ is given by,
\begin{align*}
	P &= \begin{bmatrix} 
		CB   & 0   & ... & 0 \\
		CAB  & CB  & ... & 0 \\
		\vdots & & \ddots & \vdots \\
		CA^{N-1}B & CA^{N-2}B & ... & CB
	\end{bmatrix} \in \mathbb{R}^{2N \times 2N}, \\
	A &= \begin{bmatrix}
		A_{\alpha} & 0 \\ 0 & A_{\beta}
	\end{bmatrix},\, 
	B = \begin{bmatrix}
		B_{\alpha} & 0 \\ 0 & B_{\beta}
	\end{bmatrix},\, 
	C = \begin{bmatrix}
		C_{\alpha} & 0 \\ 
		0 & C_{\beta}
	\end{bmatrix}, 
\end{align*}
with $A_{i} \in \mathbb{R}^{2 \times 2}$, $B_{i} \in \mathbb{R}^{2 \times 1}$, $C_{i} \in \mathbb{R}^{1 \times 2}\,,i = \alpha, \beta$, representing the state space representation of the closed loop dynamics of $\alpha$ and $\beta$, discretized using exact discretization. The (extended) cost function (as used similarly in \cite{mhofer:DesFabModCtrl_V3}) is given by,
\begin{multline}
	J^{j+1}(u^{j+1}) = \frac{1}{2} \biggr[{e^{j+1}}^T W_e  e^{j+1} +\nonumber \\ (u^{j+1} - u^{j})^T W_{\Delta u} (u^{j+1} - u^{j}) + {u^{j+1}}^T D^T W_{\dot{u}} D u^{j+1}\biggr]\,,  \label{eq:ILC_extended_cost}
\end{multline} 
where $W_e \succeq 0 , W_{\Delta u} \succ 0, W_{\dot{u}} \succeq 0$ in  $\mathbb{R}^{2N \times 2N}$ represent the (fixed) tuning/weighting matrices. The matrix $D$ corresponds to a discrete-time first-order derivative operator,
\begin{align}
	D = \frac{1}{T_s} \tilde{D} \otimes I,
\end{align}
where $I \in \mathbb{R}^{2 \times 2}$ represents the identity matrix, $T_s$ the sampling time, $\otimes$ the Kronecker product, and 
\begin{align}
	\tilde{D} = \begin{bmatrix} 
		-1 & 1 & ... & 0 & 0 \\
		0 & -1 & ... & 0 & 0 \\
		\vdots & \vdots & \ddots & \vdots & \vdots \\
		0 & 0 & ... & -1 & 1 \\
		0 & 0 & ... & 0 & 0
	\end{bmatrix} \in \mathbb{R}^{N \times N}.
\end{align} 
The terms in the cost function can be interpreted as follows. The first term penalizes the error in the next iteration, the second term penalizes changes in the reference correction signal from iteration to iteration and the last term allows us to restrict fast changes in the correction signal. As no constraints with respect to the correction signal are considered, the optimal solution can be computed in closed form,
\begin{align}
	{u^{j + 1}}^* &= \arg \min \{ J^{j+1}(u^{j+1}) \} \\
	&= Q u^j + L e^j, \\
	Q &= (P^T W_e P + W_{\Delta u}  + D^T W_{\dot{u}} D)^{-1} ( P^T W_e P + W_{\Delta u} ) \nonumber \\
	L&= (P^T W_e P + W_{\Delta u}  + D^T W_{\dot{u}} D)^{-1} P^T W_e. \nonumber 
\end{align}
The experimental results when applying the ILC scheme are shown in Fig. \ref{fig:ILC_noMass_Mass} with and without a mass attached. The proposed ILC approach allows for accurate reference tracking of aggressive trajectories independent of the attached load mass, while using the same tuning matrices. For instance, with a mass attached, the root-mean-square error in $\alpha$ can be reduced from 3.3$^\circ$ in the first iteration (no learning) to 0.32$^\circ$ in iteration 24, while the maximum norm of the error can be reduced from 18.9$^\circ$ to 1.4$^\circ$. Similar improvements are obtained for the case with no mass attached, illustrating the advantage of the serial architecture of the ILC.

\section{Application}
\label{sec:Application}
In this section, the pick-and-place application is described. First, the procedure to axially approach and pick an object by means of the axial compliance of the soft joint is presented. Subsequently, the experimental results and the procedure for the realization of the pick-and-place operation are discussed.  

\subsection{Picking an Object}
\begin{figure}
	\centering
	\includegraphics[scale=\pythonfigwidth]{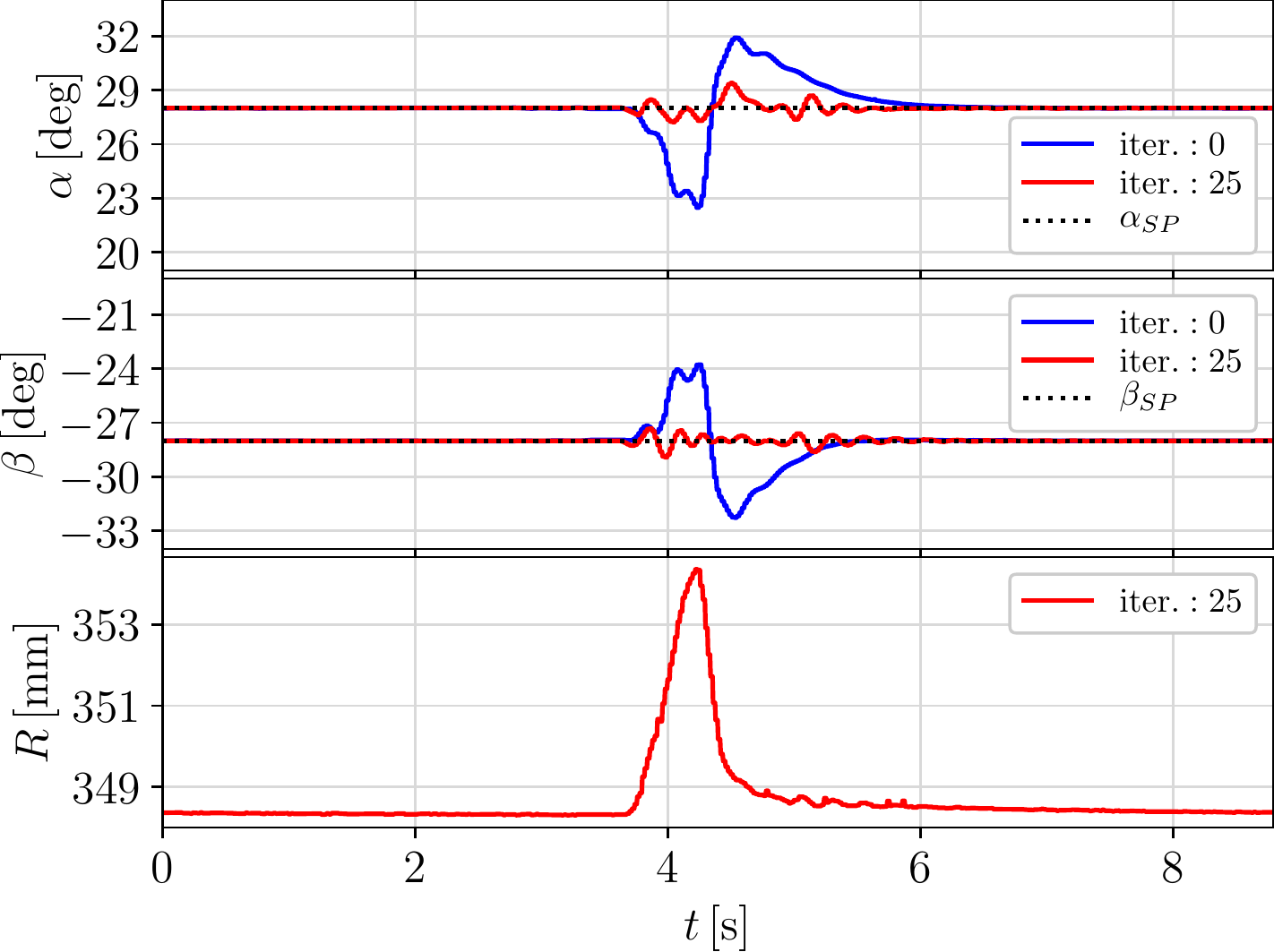}
	\vspace{\spacetocaption}
	\caption{Experimental results when the robot arm moves in the axial direction. Application of ILC allows for constant reference tracking with respect to the orientations $\alpha, \beta$, while simultaneously expanding in the axial direction. The axial deformation is induced by a rapid increase in $\bar{p}$ causing the radius of the robot arm to increase from $R_0 = \unit[347.9]{mm}$ to \unit[353.9]{mm} (bottom plot). The ILC scheme allows us to reduce the maximum error in $\alpha$ from 5.5$^\circ$ to 1.4$^\circ$ and in $\beta$ from 4.2$^\circ$ to 0.9$^\circ$ after 25 iterations of learning.}
	\label{fig:pickObj_ILC_alpha_beta_radius}
	\vspace{\spacetofig}
\end{figure}

In order to pick an object in a robust and reliable way with a vacuum gripper, it is crucial to axially approach and adhere to the object. As the system is equipped with a soft joint (see Sec. \ref{sec:SoftRoboticArm}) with certain axial compliance, objects can be approached in the axial direction. To adjust the axial elongation of the soft joint, the lower pressure bound $\bar{p}$ is increased (see Fig. \ref{fig:ILC_all_in_one_cmprsn}, bottom). However, for rapid increases in $\bar{p}$ this causes an error with respect to the constant angle reference which cannot be compensated by the feedback controller. Note that the Delta Representation solely ensures a decoupling between $\alpha, \beta$, but not their invariance with respect to $\bar{p}$. In order to compensate for the resulting error with respect to the constant angle reference, the same ILC strategy is applied when picking an object as introduced in the previous section. Note that the scope of the underlying model is restricted to quasi static changes with respect to $\bar{p}$. However, the model is sufficiently descriptive for the ILC synthesis. The ability of the system to track constant references while simultaneously moving in the axial direction using ILC is shown in Fig. \ref{fig:pickObj_ILC_alpha_beta_radius}. Note that currently there is no feedback control with respect to the axial elongation, simply relying on the feedforward adjustment of $\bar{p}$ to adjust the radius of the system.       

%

\subsection{Pick-and-Place}
\begin{figure}
	\centering
	\includegraphics[scale=\pythonfigwidth]{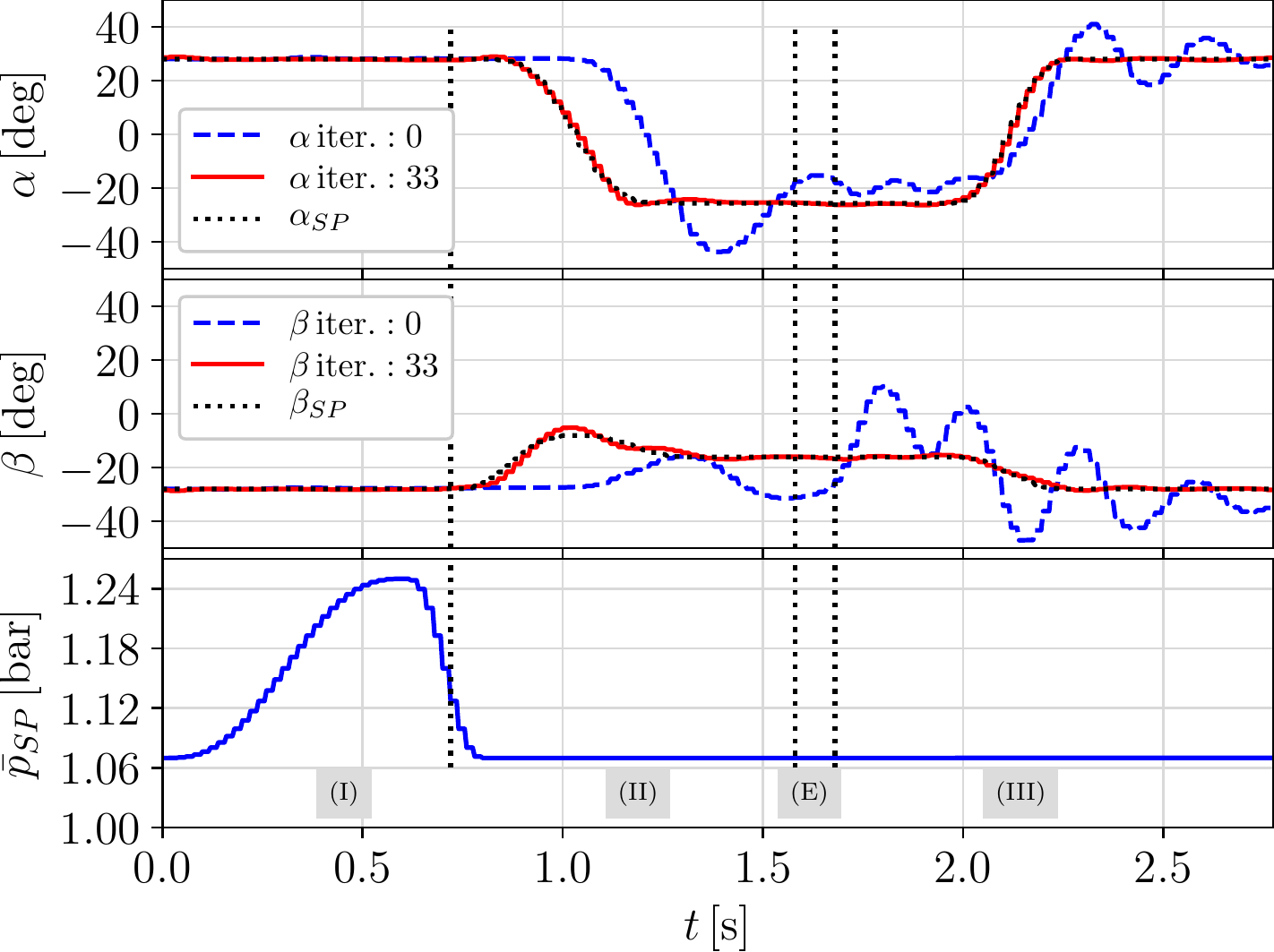}
	\vspace{\spacetocaption}
	\caption{Experimental results for an aggressive pick-and-place operation, when the warm start is applied (iteration 0) and after applying the ILC scheme for 33 additional iterations on all phases combined. The area marked as (E) corresponds to the time where the ejection impulse is switched on in order to ensure rapid purging of the vacuum when depositing an object. In the first iteration, large oscillations in $\beta$ are induced during and after phase (E) when depositing the object due to the significant change of mass in the system. Note the ability of the ILC to compensate for this effect due to its non-causal nature. The maximum errors during phase (E) can be reduced to approximately $0.5^\circ$ in both $\alpha$ and $\beta$, while depositing an object of comparable weight to the movable link of the robotic arm. The value of $\bar{p}$ is kept constant during phase (II) and (III). }
	\label{fig:ILC_all_in_one_cmprsn}
	\vspace{\spacetofig}
\end{figure}
The pick-and-place application consists of three phases, namely (I) picking an object, (II) carrying a load from the initial to the terminal position and (III) returning to the initial position after depositing the object into a box. The full duration of a pick-and-place period is \unit[2.78]{s} for a reference change of approximately $60^\circ$. The reference change occurs within \unit[0.6]{s} in phase (II) and within \unit[0.3]{s} in phase (III). An ejection impulse is applied between phases (II) and (III) when depositing an object to ensure rapid purging of the vacuum by the active pressurization of the suction cup (see phase (E) in Fig. \ref{fig:ILC_all_in_one_cmprsn}). 

In order to track the pick-and-place trajectories accurately, the ILC strategy from the previous section is applied. This, however, requires the ILC to be initialized with a warm start to ensure that the deposition of the object does not occur in an uncontrolled way due to the large overshoots when no learning is applied in aggressive maneuvers (see Fig. \ref{fig:ILC_noMass_Mass}). To this end, all three phases of the pick-and-place application are first trained independently. The experimental results of the independently trained task are depicted in Fig. \ref{fig:pickObj_ILC_alpha_beta_radius} for phase (I), Fig. \ref{fig:ILC_noMass_Mass} (bottom) for phase (II), and Fig. \ref{fig:ILC_noMass_Mass} (top) for the phase (III). The learned correction signals are truncated appropriately and concatenated resulting in the warm start correction signal. Subsequently, the ILC for the pick-and-place application is trained for the entire pick-and-place operation jointly, initialized with the warm start. The mass parameter for the underlying feedback control loop is changed continuously between the different phases. Fig. \ref{fig:ILC_all_in_one_cmprsn} shows the tracking performance during the pick-and-place application when applying the warm start and after the ILC scheme applied to the entire application has converged.

The implemented ILC approach allows for accurate and fast pick-and-place tasks. Note the high accuracy when depositing the mass, which is attributable to the non-causal nature of the ILC scheme. The interested reader is referred to Fig. \ref{fig:im_sequence_pp} and to the video attachment to gain a visual impression of the system during operation and the application performed (\url{https://youtu.be/0ovIZ-R81sg}). After convergence of the ILC, the pick-and-place application works reliably, which is demonstrated in the video attachment with 50 out of 50 consecutive successful trials.    

\newcommand{\scalePPfig}{0.062}
\newcommand{\scalePPmp}{0.24}
\begin{figure}
\centering 
\begin{minipage}{\scalePPmp\columnwidth}
\begin{tikzpicture}
	\node[inner sep = 0pt] at (0,0) {\includegraphics[scale=\scalePPfig]{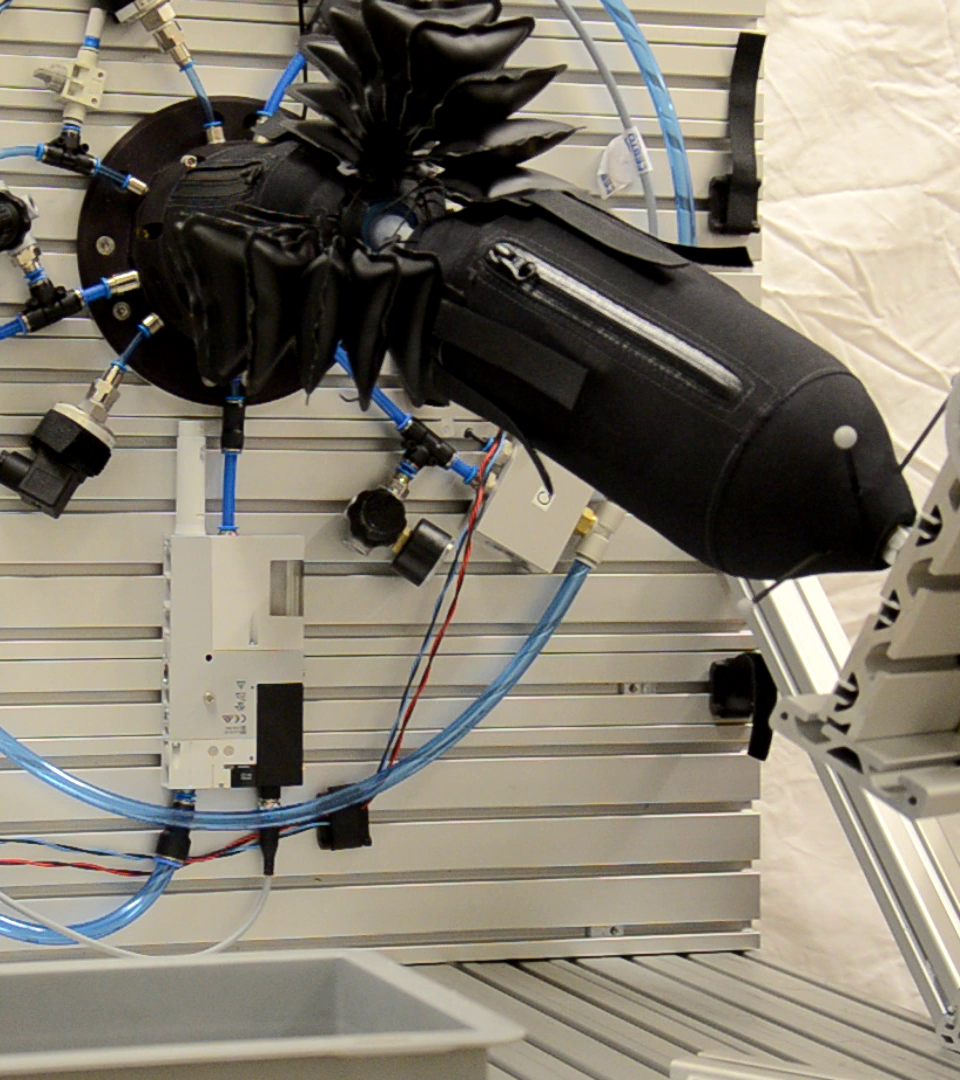}};
	\draw [fill = white, opacity=0.8, draw = none] (0.1,-0.6) rectangle (0.95,-1.1) node[pos=0.5, text=black] {\unit[0.00]{s}};
\end{tikzpicture}
\end{minipage}
\begin{minipage}{\scalePPmp\columnwidth}
\begin{tikzpicture}
	\node[inner sep = 0pt] at (0,0) {\includegraphics[scale=\scalePPfig]{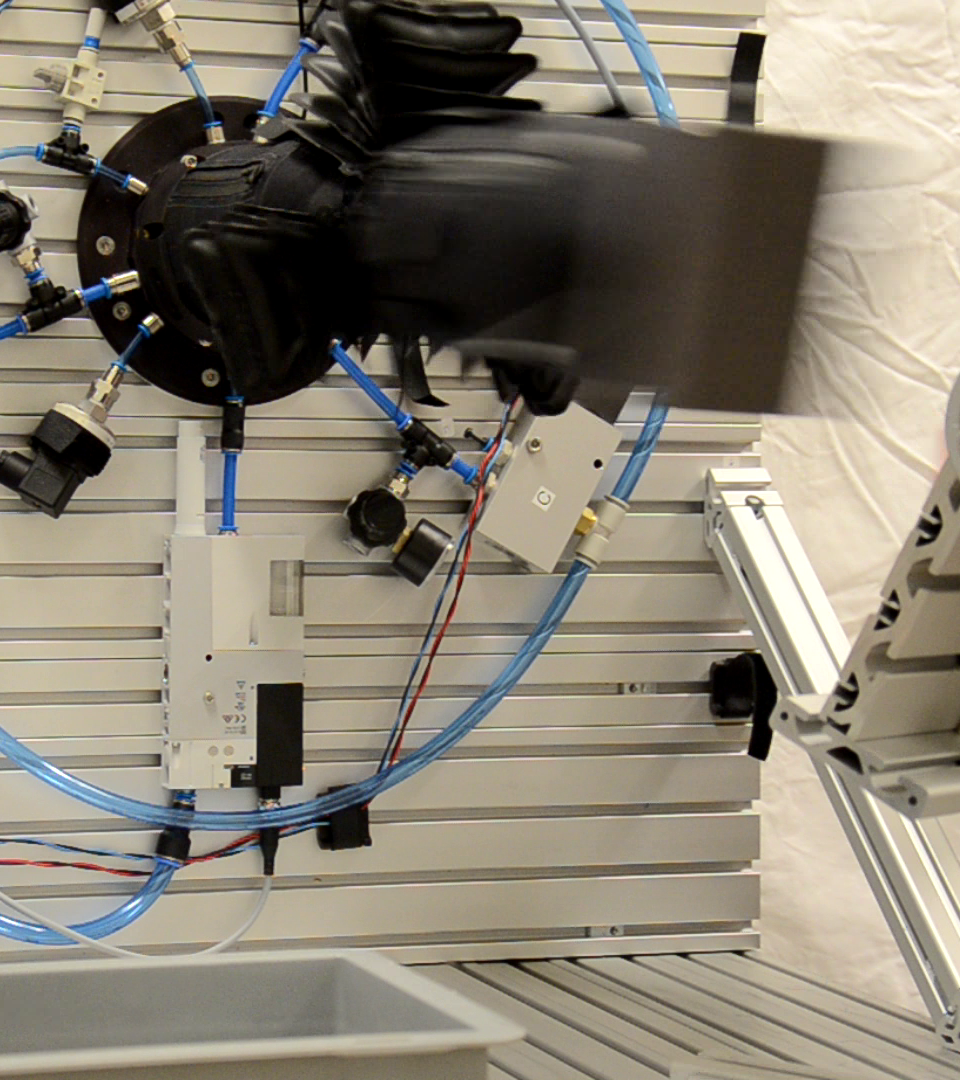}};
	\draw [fill = white, opacity=0.8, draw = none] (0.1,-0.6) rectangle (0.95,-1.1) node[pos=0.5, text=black] {\unit[1.02]{s}};
\end{tikzpicture}
\end{minipage}
\begin{minipage}{\scalePPmp\columnwidth}
\begin{tikzpicture}
	\node[inner sep = 0pt] at (0,0) {\includegraphics[scale=\scalePPfig]{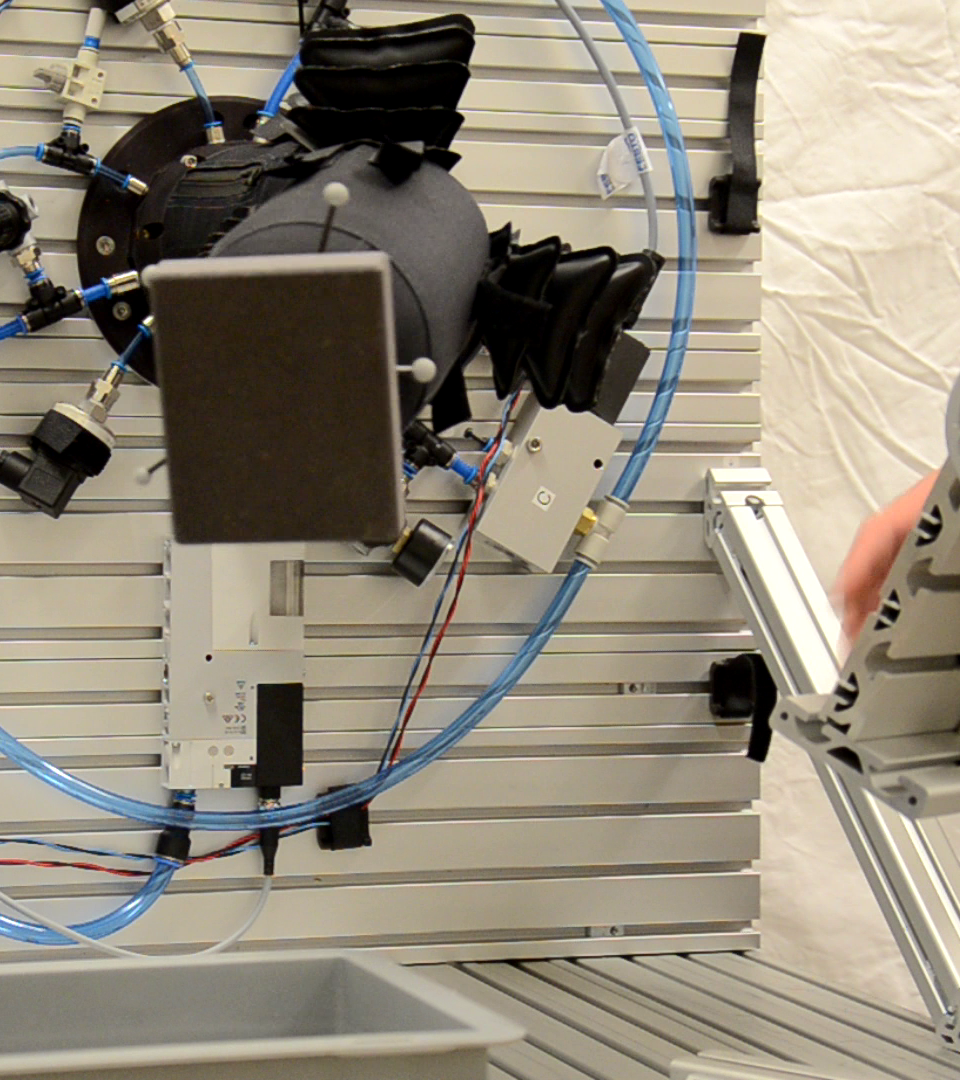}};
	\draw [fill = white, opacity=0.8, draw = none] (0.1,-0.6) rectangle (0.95,-1.1) node[pos=0.5, text=black] {\unit[1.57]{s}};
\end{tikzpicture}	
\end{minipage}
\begin{minipage}{\scalePPmp\columnwidth}
\begin{tikzpicture}
	\node[inner sep = 0pt] at (0,0) {\includegraphics[scale=\scalePPfig]{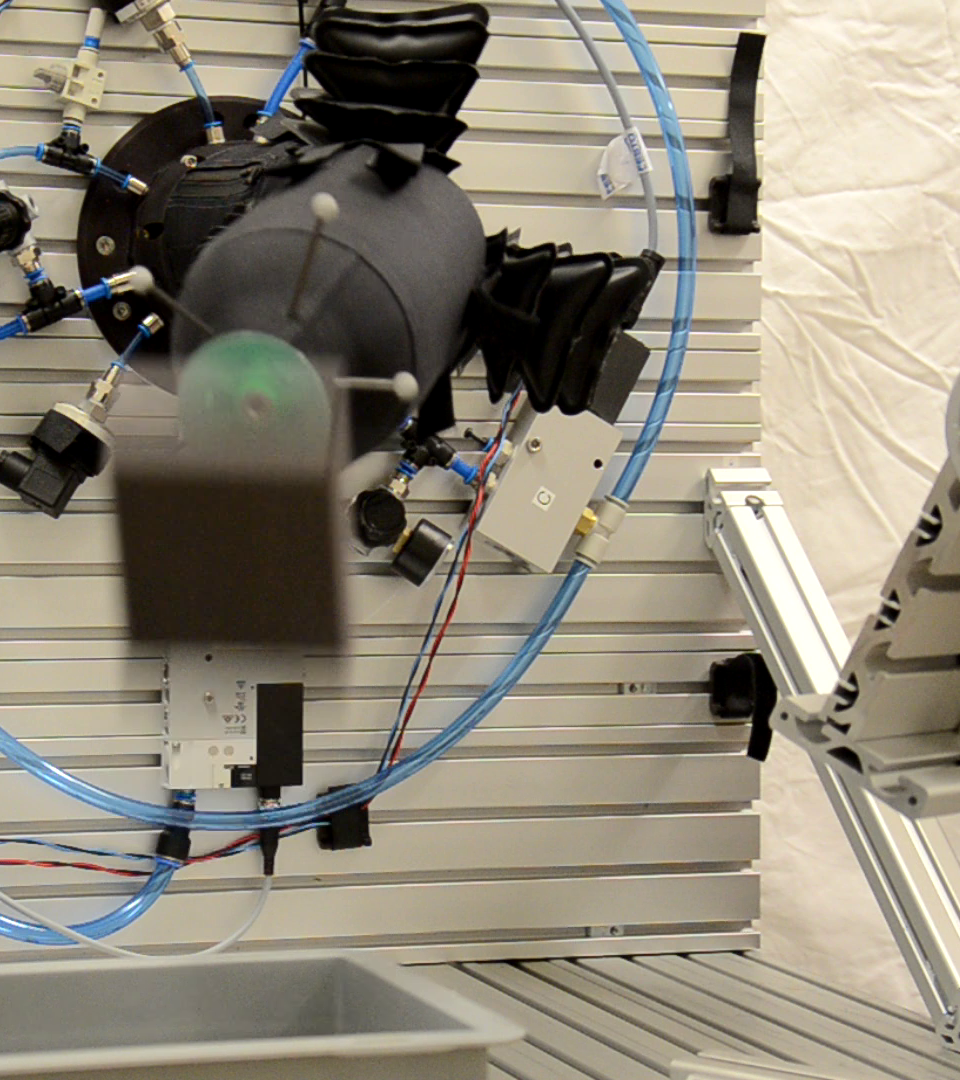}};
	\draw [fill = white, opacity=0.8, draw = none] (0.1,-0.6) rectangle (0.95,-1.1) node[pos=0.5, text=black] {\unit[1.89]{s}};
\end{tikzpicture}	
\end{minipage}
	\caption{Image sequence of the pick-and-place application. Starting with the picking of an object, the transition from the initial to the terminal position and the deposition of the object.}
	\label{fig:im_sequence_pp}
	\vspace{\spacetofig}
\end{figure}
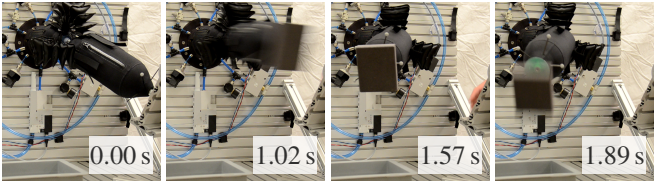

\section{Conclusion}\label{sec:Conclusion}
A reliable pick-and-place application is successfully implemented on a soft robotic arm. It is shown that one translational/stiffness and two rotational degrees of freedom can be controlled simultaneously using a soft joint. A control allocation based on pressure differences and the antagonistic actuation principle is introduced, allowing us to significantly reduce the complexity of the developed LPV model due to its decoupling property. Based on this model, a gain-scheduled feedback controller is designed, which shows comparable tracking performance irrespective of the load mass attached. An ILC scheme enables accurate tracking during aggressive maneuvers.

Currently, it is assumed that the mass is known to the feedback controller. Future work will address the estimation of the load mass online based on the model presented, the consideration of variable stiffness along the trajectories and the manipulation of different types of objects.     

\section*{Acknowledgment}

The authors would like to thank B. Walser, H. Yaohui, W. Bachmann, H. Hanimann, and the Niederberger-Kobold Foundation for their contributions to this work.

\bibliographystyle{IEEEtran}
\bibliography{bibliography}

\begin{thebibliography}{10}
\providecommand{\url}[1]{#1}
\csname url@samestyle\endcsname
\providecommand{\newblock}{\relax}
\providecommand{\bibinfo}[2]{#2}
\providecommand{\BIBentrySTDinterwordspacing}{\spaceskip=0pt\relax}
\providecommand{\BIBentryALTinterwordstretchfactor}{4}
\providecommand{\BIBentryALTinterwordspacing}{\spaceskip=\fontdimen2\font plus
\BIBentryALTinterwordstretchfactor\fontdimen3\font minus
  \fontdimen4\font\relax}
\providecommand{\BIBforeignlanguage}[2]{{%
\expandafter\ifx\csname l@#1\endcsname\relax
\typeout{** WARNING: IEEEtran.bst: No hyphenation pattern has been}%
\typeout{** loaded for the language `#1'. Using the pattern for}%
\typeout{** the default language instead.}%
\else
\language=\csname l@#1\endcsname
\fi
#2}}
\providecommand{\BIBdecl}{\relax}
\BIBdecl

\bibitem{rus2015overview}
D.~Rus and M.~T. Tolley, ``Design, fabrication and control of soft robots,''
  \emph{Nature}, vol. 521, no. 7553, 2015.

\bibitem{polygerinos_overview}
P.~Polygerinos, N.~Correll, S.~A. Morin, B.~Mosadegh, C.~D. Onal, K.~Petersen,
  M.~Cianchetti, M.~T. Tolley, and R.~F. Shepherd, ``Soft robotics: Review of
  fluid-driven intrinsically soft devices; manufacturing, sensing, control, and
  applications in human-robot interaction,'' \emph{Advanced Engineering
  Materials}, vol.~19, no.~12, 2017.

\bibitem{gaiser2012overview}
I.~Gaiser, R.~Wiegand, O.~Ivlev, A.~Andres, H.~Breitwieser, S.~Schulz, and
  G.~Bretthauer, \emph{\BIBforeignlanguage{english}{Compliant Robotics and
  Automation with Flexible Fluidic Actuators and Inflatable Structures}}.\hskip
  1em plus 0.5em minus 0.4em\relax {InTech, New York}, 2012.

\bibitem{sanan_pHRI_soft}
S.~{Sanan}, M.~H. {Ornstein}, and C.~G. {Atkeson}, ``Physical human interaction
  for an inflatable manipulator,'' in \emph{Annual International Conference of
  the IEEE Engineering in Medicine and Biology Society}, 2011.

\bibitem{abidi_onIntrinsicSafety}
H.~Abidi and M.~Cianchetti, ``On intrinsic safety of soft robots,''
  \emph{Frontiers in Robotics and AI}, vol.~4, 2017.

\bibitem{polygerRapidSoft}
B.~Mosadegh, P.~Polygerinos, C.~Keplinger, S.~Wennstedt, R.~F. Shepherd,
  U.~Gupta, J.~Shim, K.~Bertoldi, C.~J. Walsh, and G.~M. Whitesides,
  ``Pneumatic networks for soft robotics that actuate rapidly,'' \emph{Advanced
  functional materials}, vol.~24, no.~15, 2014.

\bibitem{mhofer:ILCaggr}
M.~{Hofer}, L.~{Spannagl}, and R.~{D’Andrea}, ``Iterative learning control
  for fast and accurate position tracking with an articulated soft robotic
  arm,'' in \emph{IEEE/RSJ International Conference on Intelligent Robots and
  Systems (IROS)}, 2019.

\bibitem{althoefer2018antagonistic}
K.~Althoefer, ``Antagonistic actuation and stiffness control in soft inflatable
  robots,'' \emph{Nature Reviews Materials}, vol.~3, no.~6, 2018.

\bibitem{gillespie:simPosStiffCtrl}
M.~T. {Gillespie}, C.~M. {Best}, and M.~D. {Killpack}, ``Simultaneous position
  and stiffness control for an inflatable soft robot,'' in \emph{IEEE
  International Conference on Robotics and Automation (ICRA)}, 2016.

\bibitem{mhofer:DesFabModCtrl_V3}
M.~Hofer and R.~D’Andrea, ``Design, fabrication, modeling and control of a
  fabric-based spherical robotic arm,'' \emph{Mechatronics}, vol.~68, 2020.

\bibitem{LPV_modCtrlHinf}
Z.~{Qiao}, P.~H. {Nguyen}, P.~{Polygerinos}, and W.~{Zhang}, ``Dynamic modeling
  and motion control of a soft robotic arm segment,'' in \emph{2019 American
  Control Conference (ACC)}, 2019.

\bibitem{angeliniILCsoft_decentra}
F.~{Angelini}, C.~D. {Santina}, M.~{Garabini}, M.~{Bianchi}, G.~M. {Gasparri},
  G.~{Grioli}, M.~G. {Catalano}, and A.~{Bicchi}, ``Decentralized trajectory
  tracking control for soft robots interacting with the environment,''
  \emph{IEEE Transactions on Robotics}, vol.~34, no.~4, 2018.

\bibitem{katzschmann_graspAndPlace}
R.~K. Katzschmann, A.~D. Marchese, and D.~Rus, ``Autonomous object manipulation
  using a soft planar grasping manipulator,'' \emph{Soft robotics}, vol.~2,
  no.~4, 2015.

\bibitem{hyattPP_space}
P.~Hyatt, D.~Kraus, V.~Sherrod, L.~Rupert, N.~Day, and M.~D. Killpack,
  ``Configuration estimation for accurate position control of large-scale soft
  robots,'' \emph{IEEE/ASME Transactions on Mechatronics}, vol.~24, no.~1,
  2018.

\bibitem{rusILCsoft_grasping}
A.~D. Marchese, R.~Tedrake, and D.~Rus, ``Dynamics and trajectory optimization
  for a soft spatial fluidic elastomer manipulator,'' \emph{The International
  Journal of Robotics Research}, vol.~35, no.~8, 2016.

\bibitem{festoBionicHandling}
A.~Grzesiak, R.~Becker, and A.~Verl, ``The bionic handling assistant: A success
  story of additive manufacturing,'' \emph{Assembly Automation}, vol.~31, 09
  2011.

\bibitem{polygerSoftPolyLimb}
P.~H. Nguyen, C.~Sparks, S.~G. Nuthi, N.~M. Vale, and P.~Polygerinos, ``Soft
  poly-limbs: Toward a new paradigm of mobile manipulation for daily living
  tasks,'' \emph{Soft robotics}, vol.~6, no.~1, 2019.

\bibitem{mhofer:DesModCtrl_V2}
M.~{Hofer} and R.~{D'Andrea}, ``Design, modeling and control of a soft robotic
  arm,'' in \emph{IEEE/RSJ International Conference on Intelligent Robots and
  Systems (IROS)}, 2018.

\bibitem{zughaibi_onlineAppx}
J.~Zughaibi, M.~Hofer, and R.~D'Andrea, ``Online appendix - {A} fast and
  reliable pick-and-place application with a spherical soft robotic arm,''
  Available: \url{https://doi.org/10.3929/ethz-b-000470203}, 2021.

\bibitem{ljung_sysID}
L.~Ljung, \emph{System Identification (2nd Ed.): Theory for the User}.\hskip
  1em plus 0.5em minus 0.4em\relax USA: Prentice Hall PTR, 1999.

\bibitem{ReductionComplianceIssue}
C.~{Della Santina}, M.~{Bianchi}, G.~{Grioli}, F.~{Angelini}, M.~{Catalano},
  M.~{Garabini}, and A.~{Bicchi}, ``Controlling soft robots: Balancing feedback
  and feedforward elements,'' \emph{IEEE Robotics Automation Magazine},
  vol.~24, no.~3, 2017.

\bibitem{briatLPVtimeDelaySys}
C.~Briat, ``Linear parameter-varying and time-delay systems,'' \emph{Analysis,
  observation, filtering \& control}, vol.~3, 2014.

\bibitem{bristow2006survey}
D.~A. {Bristow}, M.~{Tharayil}, and A.~G. {Alleyne}, ``A survey of iterative
  learning control,'' \emph{IEEE Control Systems Magazine}, vol.~26, no.~3,
  2006.

\bibitem{owensILCoptimization}
D.~H. Owens and J.~H{\"a}t{\"o}nen, ``Iterative learning control — an
  optimization paradigm,'' \emph{Annual reviews in control}, vol.~29, no.~1,
  2005.

\end{thebibliography}

\end{document}